%% file: main.tex
\documentclass[sigconf,authorversion,nonacm]{aamas}

\usepackage{bm}
\usepackage{balance} %

\acmSubmissionID{27}

\title[Active Third-Person Imitation Learning]{Active Third-Person Imitation Learning}

\author{Timo Klein}
\affiliation{
  \institution{University of Vienna}
  \institution{Faculty of Computer Science}
  \institution{Doctoral School Computer Science}
  \city{Vienna}
  \country{Austria}}
\email{timo.klein@univie.ac.at}

\author{Susanna Weinberger}
\affiliation{
  \institution{University of Vienna}
  \institution{Faculty of Computer Science}
  \city{Vienna}
  \country{Austria}}

\author{Adish Singla}
\affiliation{
  \institution{Max Planck Institute of Software Systems}
  \city{Saarbrücken}
  \country{Germany}}
  \email{adishs@mpi-sws.org}

\author{Sebastian Tschiatschek}
\affiliation{
  \institution{University of Vienna}
  \institution{Faculty of Computer Science}
  \city{Vienna}
  \country{Austria}}
  \email{sebastian.tschiatschek@univie.ac.at}

\input{0_abstract}

\keywords{Imitation Learning, Inverse Reinforcement Learning, Active Learning}

\newcommand{\BibTeX}{\rm B\kern-.05em{\sc i\kern-.025em b}\kern-.08em\TeX}

\input{macros.tex}

\begin{document}

\pagestyle{fancy}
\fancyhead{}

\maketitle

\input{1_introduction}

\input{2_problem-setting}

\input{5_related}

\input{3_theory}

\input{4_methodology}

\input{6_experiments}
  
\input{7_conclusion}

\begin{acks}
We thank Lukas Miklautz and Simon Rittel for their valuable feedback on earlier versions of this work. We thank Kevin Sidak for the insightful discussions. Lastly, we want to thank the open source communities of NumPy \cite{harrisArrayProgrammingNumPy2020}, Weights \& Biases \cite{biewaldExperimentTrackingWeights2020a}, plotly \cite{plotly-communityPlotlyPlotlyPy} and Pytorch \cite{paszkePyTorchImperativeStyle} for providing the tools used in this study.
\end{acks}

\clearpage
\bibliographystyle{ACM-Reference-Format} 
\bibliography{refs}

\clearpage

\appendix

\input{a8_appendix-proofs}

\input{a9_additional_results}

\input{a10_env_descriptions}

\input{a11_correlation_selection}

\input{a12_gail_background}

\input{a13_extended_related_work}

\input{a14_compute_parameters}

\input{a18_illustration_thm}

\end{document}

%% file: 0_abstract.tex
\begin{abstract}
We consider the problem of third-person imitation learning with the additional challenge that the learner must select the perspective from which they observe the expert. In our setting, each perspective provides only limited information about the expert's behavior, and the learning agent must carefully select and combine information from different perspectives to achieve competitive performance. This setting is inspired by real-world imitation learning applications, e.g., in robotics, a robot might observe a human demonstrator via camera and receive information from different perspectives depending on the camera's position. We formalize the aforementioned active third-person imitation learning problem, theoretically analyze its characteristics, and propose a generative adversarial network-based active learning approach. Empirically, we demonstrate that our proposed approach can effectively learn from expert demonstrations and explore the importance of different architectural choices for the learner's performance.

\end{abstract}

%% file: macros.tex
\usepackage{lipsum}
\usepackage{subcaption}
\usepackage[inline]{enumitem}

\usepackage{amsmath,amsthm}
\usepackage{mathtools}
\usepackage{bm}
\usepackage{xfrac}

\usepackage{tikz}
\usetikzlibrary{calc}
\usetikzlibrary{fit}
\usetikzlibrary{positioning}
\usetikzlibrary{arrows.meta}

\usepackage{fontawesome}

\usepackage[toc,page,header]{appendix}
\usepackage{minitoc}

\usepackage{algpseudocode}
\usepackage[algo2e,linesnumbered, ruled, vlined,noend]{algorithm2e}
\usepackage{algorithm}

\newtheorem{theorem}[]{Theorem}
\newtheorem{observation}[]{Observation}

\DeclareMathOperator*{\argmin}{argmin}
\algnewcommand\algorithmicforeach{\textbf{for each:}}
\algnewcommand\ForEach{\item[ \algorithmicforeach]}

\usepackage{subcaption}

\usepackage{xcolor}
\definecolor{codegreen}{rgb}{0,0.6,0}
\definecolor{codegray}{rgb}{0.5,0.5,0.5}
\definecolor{codepurple}{rgb}{0.58,0,0.82}
\definecolor{backcolour}{rgb}{0.95,0.95,0.92}
\definecolor{colororange}{HTML}{FF6900}
\definecolor{colorpurple}{HTML}{80368C}
\definecolor{colorturquise}{HTML}{95D1BD}
\definecolor{colormyblue}{HTML}{0077B5}
\definecolor{colorpink}{HTML}{DD2A7B}
\definecolor{mygreen}{RGB}{0, 173, 81}

\usepackage{ifthen}
\newboolean{include-notes}
\setboolean{include-notes}{true}
\newcommand{\nb}[3]{\ifthenelse{\boolean{include-notes}}{{\colorbox{#2}{\bfseries\sffamily\scriptsize\textcolor{white}{#1}}}{\ \textcolor{#2}{\sf\small\textit{#3}}}}{}}

\newcommand*{\transpose}{\ensuremath{{\mathsf T}}}
\newcommand*{\matvec}[1]{\ensuremath\bm{#1}}

\DeclareMathOperator{\sign}{Sign}
\newcommand*{\environment}{\ensuremath{e}}
\newcommand*{\perspective}{\ensuremath{\nu}}
\newcommand*{\perspectives}{\ensuremath{\mathcal{V}}}
\newcommand*{\strategy}{\ensuremath{\xi}}

\newcommand*{\feats}{\ensuremath\matvec{\phi}(s, a)}
\newcommand*{\featspersp}{\ensuremath\matvec{\phi}_\nu (s, a)}
\newcommand*{\linearpersp}{\ensuremath\matvec{A}_\nu}
\newcommand*{\cumfeats}{\ensuremath\matvec{\Psi}}

\newcommand*{\actions}{\ensuremath{\mathcal{A}}}
\newcommand*{\horizon}{\ensuremath{H}}
\newcommand*{\mdp}{\ensuremath{\mathcal{M}}}
\newcommand*{\observations}{\ensuremath{\mathcal{O}}}
\newcommand*{\states}{\ensuremath{\mathcal{S}}}
\newcommand*{\transitions}{\ensuremath{\mathcal{T}}}

\newcommand*{\discriminator}{\ensuremath{\mathbf{D}}}
\newcommand*{\expert}{\ensuremath{\textnormal{E}}}
\newcommand*{\learner}{\ensuremath{\textnormal{L}}}
\newcommand*{\CE}{\ensuremath{\textnormal{CE}}}

%% file: 1_introduction.tex
\section{Introduction}
\begin{figure}
    \centering
    \scalebox{1.00}{\input{tikz/scheme.tex}}
    \caption{Schema of the \emph{active third-person imitation learning} problem.
    The learner can select the perspective, e.g., the angle of a camera in a 3D setting, from which it observes the expert's demonstration. Effective learning requires the selection of informative perspectives using a perspective selection strategy $\strategy$ while avoiding uninformative ones.}
    \label{fig:schematic}
\end{figure}
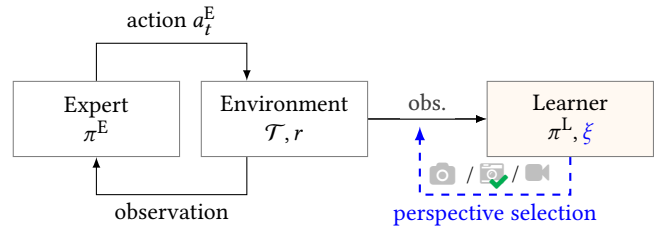
Reinforcement Learning (RL) has achieved remarkable successes in game playing~\cite{silver2017mastering,vinyals2019grandmaster,schrittwieser2020mastering}, robotics \cite{singh2022reinforcement,zhao2020sim} and industrial control \cite{coal_plant_control, power_grid_congestion_management}. However, its need for well-defined reward functions is often its Achilles heel: Specifying reward functions usually requires significant domain expertise and effort. 
One option enabling agents to learn in such settings, Learning from Demonstrations (LfD), is a well-established alternative~\cite{hussein2017imitation,arora2021survey,abbeel2004apprenticeship}. In LfD, a learning agent observes demonstrations from an expert to infer rewarding behavior. Typically, it is assumed that the expert's demonstrations are close to optimal and that the learner can optimize the expert's implicit reward function by mimicking its behavior. Two common approaches for LfD are Inverse RL (IRL)~\cite{ng2000algorithms,russell_learning_1998} and Imitation Learning (IL)~\cite{hussein2017imitation}. In this paper, we focus on the latter. The classical formulation of IL assumes that the learner can observe state-action trajectories of the expert's behavior~\cite{ross2010efficient}. The learning agent's policy can then be optimized via supervised learning by predicting the expert's actions for given states. Yet, in many practical scenarios, e.g., when learning from a human demonstrator through cameras, observing expert actions and the environment's ground-truth states is infeasible. Thus, an agent must learn from observations without having access to actions~\cite{torabi_recent_2019}.

But there is a further---commonly neglected---challenge, namely from which perspective the learner should observe the demonstrator to learn efficiently. To make this more concrete, consider the following motivating example: A human wants to teach a kitchen robot how to chop vegetables. Specifying a reward function that assigns rewards for each low-level behavior and individual vegetable is tedious if not infeasible. When applying IL to learn from demonstrations, the kitchen robot must select the perspective from which to watch the demonstrator. Watching the expert chop from the side reveals information about correctly moving the knife, whereas watching from the front discloses crucial details on the positioning of the fingers. Choosing a perspective that does not show the demonstrator does not support learning.
This example highlights that
\begin{enumerate*}[label=(\roman*)]
    \item information from different perspectives can be complementary and
    \item that a learner must actively and deliberately choose the perspectives to learn efficiently.
\end{enumerate*}
These features distinguish our work from other related works like~\cite{sermanetTimeContrastiveNetworksSelfSupervised2018,shangSelfSupervisedDisentangledRepresentation2021,garelloThirdPersonVisualImitation2022,zakkaXIRLCrossembodimentInverse} that perform imitation learning from multiple perspectives but involve no need to actively choose good perspectives, cf.\ Section~\ref{sec:related} for details.
Other settings in which actively choosing the perspective for imitation learning can be important include applications with power and privacy constraints and applications with different perspectives that are expensive to provide, e.g., because of required human effort.
Inspired by this, we introduce the problem of \textbf{active third-person IL} in which the learner can \emph{choose among a set of perspectives} when observing the expert. Figure~\ref{fig:schematic} illustrates our considered setting and highlights differences compared to traditional third-person IL. Both settings consider a problem where the learner's observation space differs from the expert's, but only in our setting the learner can influence the perspective from which it observes the expert.

In this paper, we formalize and theoretically analyze third-person IL and characterize the limits of what can be learned regarding the available perspectives and the structure of the underlying reward function.
Inspired by previous work on third-person IL~\cite{stadie2017third}, we propose an approach for learning in the active third-person IL setting based on generative adversarial imitation learning (GAIL)~\cite{ho2016generative}.
In our approach, we assume a finite number of possible perspectives and discriminators quantifying the \emph{imitation performance} of the learner in those perspectives. The learner implements a perspective selection strategy to decide from which perspective to observe the expert. Ultimately, the goal is that the discriminators cannot distinguish the learner's and the expert's actions from any of the perspectives. 

To summarize our contributions:
\begin{enumerate*}[label=(\roman*)]
    \item We formalize the problem of active third-person IL.

    \item We analyze the characteristics of the active third-person IL problem.
    
    \item We propose multiple approaches to account for different perspectives in active third-person IL. 
    
    \item We provide proof of concept experiments with multiple scenarios and demonstrate the effectiveness of our framework in toy and benchmark environments (MuJoCo)~\cite{brockman2016openai}.
\end{enumerate*}

Our paper is structured as follows.
We formalize the active third-person IL problem in Section~\ref{sec:setting} and discuss related work in Section~\ref{sec:related}.
Then we characterize the active third-person IL problem in Section~\ref{sec:analysis} and introduce our approach in Section~\ref{sec:approach}.
We present our experiments in Section~\ref{sec:experiments} and conclude our paper in Section~\ref{sec:conclusions}.

%% file: tikz/scheme.tex
\begin{tikzpicture}
  \node[minimum height=1.cm, draw=gray] (expert) {\parbox{2cm}{\centering Expert\\$\pi^{\expert}$}};
  
  \node[minimum height=1.cm, draw=gray, right of=expert, xshift=1.5cm] (env) {\parbox{2cm}{\centering Environment\\$\transitions,r$}};
  
  \node[minimum height=1.cm, draw=gray, right of=env, xshift=2.8cm, fill=orange!05!white] (learner) {\parbox{2cm}{\centering Learner\\$\pi^{\learner}, \textcolor{blue}{\strategy}$}};
  
  \draw[->, >=latex] (expert.north) -- ++(0,0.5) -| node[pos=0.25, above] () {action $a^{\expert}_t$} ($(env.north)+(-0.5,0)$);
  
  \draw[->, >=latex] ($(env.south)+(-0.5,0)$) -- ++(0,-0.5) -| node[pos=0.25, below] () {observation} (expert);
  
  \draw[->, >=latex] (env) -- node[pos=0.5, above] (ctrl) {obs.} (learner);
  
  \draw[->, >=latex, dashed, draw=blue, thick] (learner.south) -- ++(0,-0.5) -| node[pos=0.25, below] (selection) {\textcolor{blue}{perspective selection}} ($(ctrl.south)+(-0.1,-0.1)$);
 \node[above of=selection, yshift=-4.5mm, xshift=-7mm] (persp1) {\textcolor{gray!50!white}{\faCamera}};
  \node[right of=persp1,xshift=-5mm] (persp2) {\phantom{e}/ \textcolor{gray!50!white}{\faCameraRetro}};
  \node[right of=persp2,xshift=-4mm] (persp3) {\phantom{e}/ \textcolor{gray!50!white}{\faVideoCamera}};
  \node[xshift=2.5mm,yshift=-1.2mm, fill=none, fill opacity=1., inner sep=0pt] at (persp2) () {\scalebox{0.9}{\textcolor{mygreen}{\faCheck}}};
  
  \draw[->, >=latex] (env) -- node[pos=0.5, above, fill=white, fill opacity=0.5] (ctrl) {obs.} (learner);
  
\end{tikzpicture}

%% file: 2_problem-setting.tex
\section{Problem Setting}
\label{sec:setting}

\textbf{Basic notation \& setup.} 
We consider the problem of active third-person IL in a \emph{Markov Decision Process} (MDP).
An MDP is characterized by $\mdp = (\states, \actions, \transitions, \rho, r, \horizon)$, where $\states$ is a set of states, $\actions$ a set of actions an agent can take, $\transitions\colon \states \times \actions \rightarrow [0,1]^{|\states|}$ describes the transition probabilities into the next state by taking an action in the current state, $\rho$ is the initial state distribution, $r\colon \states \times \actions \rightarrow \mathbb{R}$ the reward function, and $\horizon$ the horizon of the interaction~\cite{sutton2018reinforcement}.
An agent implements a (stochastic) policy $\pi\colon \states\times \actions \rightarrow [0,1]$, which is characterized by the probability $\pi(s,a)$ of taking action $a$ in state $s$, and describes the agent's behavior.
The standard goal of an agent in RL is to learn a policy $\pi$ that maximizes the expected return $J(\pi) = \mathbb{E}\big[ \sum_{t=0}^{H-1} r(s_t, a_t) | \pi \big]$,
where the expectation is over the randomness in the initial state distribution, the transitions, and the policy.

\textbf{Imitation learning.}
In imitation learning (IL), an agent (learner) aims to learn rewarding behavior from an expert by observing it.
The expert provides demonstrations from a (possibly stochastic) policy $\pi^{\expert}$ which (approximately) maximizes the expected return $J(\pi^{\expert})$.
Depending on the precise setting, the learner either observes the expert's states and actions or only the expert's states.
In more general settings, the learner does not observe the states directly but only observations $\observations\colon \states \rightarrow \mathbb{R}^d$ instead. 
Suppose states/observations and actions are observed. In that case, imitation learning is often considered a supervised learning problem in which imitating the expert corresponds to learning a classifier from state/observations to the expert's actions.
Such approaches typically suffer from compounding errors but can be performed without access to the environment.
In other lines of work, IL is considered as the problem of matching state-action-occupancies of the expert and the learner, and interaction with the environment is required~\cite{ho2016generative}. Consider, e.g.,~\citet{hussein2017imitation} for a survey.

\textbf{The active third-person imitation learning problem.}
In this paper, we are interested in a more general IL setting in which the learner must actively decide how it observes the expert.
This is inspired by scenarios in which the agent could for instance be a robot, observing a human's demonstrations and deciding from which angle to watch the demonstration.
Formally, we assume that the learner can select the perspective and thereby influence the observations $o$ it receives, i.e., $(s,\perspective) \mapsto o$, where $s$ is the state of the environment and $\perspective \in  \perspectives$ is the selected perspective chosen among a set of perspectives~$\perspectives$ available to the learner and $o \in \mathbb{R}^{d_\perspective}$ is a perspective dependent observation.
In this paper, we assume that the set of possible perspectives $\perspectives$ is finite and that the learner selects the perspective for a whole demonstration.
The learner's goal is to find a policy $\pi^\learner$ that imitates the expert's behavior as quickly and accurately as possible.
To achieve this, the learner must employ a suitable perspective selection strategy $\strategy$ and effectively combine the  information from the experts' demonstrations observed from the chosen perspectives.
The interaction of the learner and expert is summarized in Algorithm~\ref{alg:interaction}. In Section~\ref{sec:approach-details} we provide details of our instantiation for learning during this interaction.

\input{algorithms/algorithm-interaction.tex}

%% file: algorithms/algorithm-interaction.tex
\newcommand\mycommfont[1]{\footnotesize\ttfamily\textcolor{blue}{#1}}
\SetCommentSty{mycommfont}

\setlength{\textfloatsep}{2mm}
\begin{algorithm2e}[t]
\caption{Interaction in the active third-person imitation learning problem}\label{alg:interaction}
\SetAlgoLined
\DontPrintSemicolon
\KwIn{environment $\mathcal{M}$, expert $\expert$, learner $\learner$, learner policy $\pi^\learner$, perspectives $\perspectives = \{ \perspective_1, \ldots, \perspective_i\}$, selection strategy $\strategy$, number of demonstrations~$K$}
\For{$i=1,\ldots,K$}{
   $\perspective_i \leftarrow$ Next perspective for observing the expert $\expert$ selected by the learner $\learner$ \;
   $\omega = (o_1, o_2, \ldots) \leftarrow$ Observations of expert $\expert$'s demonstration from perspective $\perspective_i$ \;
   $\pi^\learner_i,  \strategy_i \leftarrow$ Learner $\learner$ uses  observations $\omega$ to update its policy and strategy \; %
}%
\KwResult{Optimized learner's policy $\pi^\learner$} %
\end{algorithm2e}

%% file: 5_related.tex
\section{Related Work}
\label{sec:related}

We first review approaches for IL when expert actions are available, followed by a discussion of IL methods suitable for scenarios where the expert's actions cannot be observed, and approaches for IL from multiple perspectives.

\textbf{Imitation learning from demonstrations.}
There are two main approaches for IL from demonstrations: Behavioural cloning (BC) \cite{ross2010efficient} and Inverse Reinforcement Learning (IRL) \cite{ng2000algorithms,russell_learning_1998}. In BC, learning from demonstrations is phrased as a supervised problem in which a regressor or classifier is trained to predict an expert's action based on the expert's state~\cite{ross2010efficient}. IRL \citep{russell_learning_1998} addresses IL by inferring a reward function from expert observations. While it has already proven useful in practical applications \cite{faragbehavioralcloning}, IL is prone to compounding errors and causal confusion~\cite{de2019causal}.
Generative Adversarial Imitation Learning (GAIL)~\cite{ho2016generative} is an adaption of the GAN framework \cite{NIPS2014_5ca3e9b1} for IL circumventing some issues arising from directly learning the reward function. It does so by training a model to discriminate between the expert's and a learner's state-action pairs while using the discriminator's prediction as a reward for the learner.

\textbf{Imitation learning from observations.}
The assumption of access to the expert's actions can be limiting in many applications.
Thus, learning from observations only has received considerable attention, e.g., \citet{torabi_recent_2019} approached the challenge of learning from online videos, where only observations of expert behavior are available. Using state-only observations for IL has also been discussed in ~\citet{ijspeert_movement_2002,bentivegna_humanoid_2002, VOILA}. Recent work uses image observations to extend IL's applicability to scenarios like robotics. Various approaches, e.g.,~\cite{torabi_behavioral_2018,nair_combining_2017,pathak_zero-shot_2018} are based on inverse dynamics models, which the learner can infer without any interaction with the expert~\cite{torabi_behavioral_2018}. 
Another widely used IL variant adapts the original GAIL~\cite{ho2016generative} framework to learning from observations. Using single states as inputs to the discriminator may not be sufficient to imitate the expert \cite{merel_learning_2017,henderson_optiongan_2018}. \citet{torabi_recent_2019} provides a counterexample with an environment where the goal is to run a circle clockwise. Observing the expert's state distribution does not allow the learner to distinguish clockwise from counterclockwise circles. Follow-up work has therefore used three consecutive images as observations~\cite{torabi_generative_2019}. \citet{stadie2017third} focus on viewpoint differences and use two images $(s_t,s_{t+\Delta})$ as input to the discriminator, that are shifted by $\Delta$ time steps.

\textbf{Imitation learning from different perspectives.}
IL from demonstrations typically requires the perspective of the expert and learner to be the same.
This is a limiting assumption compared to human learning, where observing a demonstrator in a third-person view is sufficient to facilitate learning. The seminal work by \citet{stadie2017third} provides a step towards this goal by focusing on viewpoint differences and using two time-shifted images $(s_t,s_{t+\Delta})$ as input to the discriminator. The agent's reward is given by a discriminator built on top of a viewpoint-agnostic representation. 
\citet{sermanetTimeContrastiveNetworksSelfSupervised2018} also aim to learn a representation that is perspective-invariant by distinguishing between temporally distant frames from the same trajectory and temporally close frames from different trajectories. Note that their approach requires collecting perfectly aligned observations from all perspectives for each trajectory, which is a constraint not applied to our framework. Compared to learning a joint but viewpoint-invariant representation, \citet{shangSelfSupervisedDisentangledRepresentation2021} use a dual autoencoder to disentangle viewpoint and state information. This results in a perspective-invariant representation of the state. This is related to training a generative model to reconstruct embeddings of third-person frames into first-person views \cite{garelloThirdPersonVisualImitation2022}. Lastly, \citet{zakkaXIRLCrossembodimentInverse} considers IL from multiple experts with different embodiments that differ from the learner's configuration. This relaxes the assumption of perfectly corresponding observations made in previous works. All the above-mentioned methods focus on the problem of learning a viewpoint-invariant representation. In comparison, our work additionally considers the question of \emph{which perspective to choose for imitation}.

%% file: 3_theory.tex
\section{Analysis and Insights}
\label{sec:analysis}

In this section, we analyze the characteristics of the active third-person IL problem.
We focus on feature-matching-based approaches that can be readily applied in cases for which the ground truth states can not be observed.
In particular, we study how the combination of reward structure and available perspectives impacts achievable performance.
To this end, we assume that state-action pairs $(s,a)$ are associated with ground truth features $\feats \in \mathbb{R}^d$ and that rewards are (possibly non-linear) functions of these features, i.e., $r(s,a) = g(\feats)$, where $g\colon \mathbb{R}^d \rightarrow \mathbb{R}$.
Because of space constraints, all proofs are deferred to the appendix.

\textbf{Linear reward functions and linear transformations.}
We start by investigating linear reward functions and perspectives that correspond to linear transformations of the (unknown) ground truth features $\feats$, i.e., $\featspersp = \linearpersp \feats$, where $\linearpersp \in \mathbb{R}^{d_\nu \times d}$.
To this end, let $\matvec{\mu}(\pi) = \mathbb{E}[\sum_{t=0}^\infty \gamma^t \matvec{\phi}(s_t,a_t) \mid \pi]$ be the expected discounted feature for policy $\pi$.
We can then make the following observation by extending previous results of \citet{haug2018teaching}.
\begin{theorem}
    \label{thm:linear-rewards-and-projections}
    Assume a linear reward function $r(s,a) = \langle \matvec{w}^*, \feats \rangle$, where $\matvec{w}^*$ are unknown reward parameters with $\|\matvec{w}^*\| \leq 1$, and perspectives characterized by linear transformations $\{\linearpersp\}_{\perspective \in \perspectives}$ of the features $\feats$.
    Assume that the learner's policy $\pi^\learner$ matches the feature expectations of the expert's policy $\pi^\expert$ with precision $\epsilon / |\perspectives|$ for all perspectives, i.e., $\|\linearpersp(\matvec{\mu}(\pi^\expert) - \matvec{\mu}(\pi^\learner))\| < \epsilon / |\perspectives|$ for all $\perspective \in \perspectives$. Then
    \begin{align*}
        | \langle \matvec{w}^*, \matvec{\mu}(\pi^\expert) - \matvec{\mu}(\pi^\learner) \rangle | < \frac{\epsilon}{\sigma(\bar{\matvec{A}})} + \rho( \bar{ \matvec{A} }; \matvec{w}^*) \ \textnormal{diam} \, \matvec{\mu}(\Pi),
    \end{align*}
    where $\bar{\matvec{A}} = [\matvec{A}_1^\transpose, \ldots, \linearpersp^\transpose]^\transpose$ is the matrix resulting from stacking all transformation matrices, $\sigma(\bar{\matvec{A}}) = \min_{\matvec{v} \perp \textnormal{ker} \bar{\matvec{A}}, \|\matvec{v}\|=1} \|\bar{\matvec{A}} \matvec{v}\|$, $\rho(\bar{\matvec{A}}; \matvec{w}^*) = \max_{\matvec{v} \in \textnormal{ker} \bar{\matvec{A}}, \|v\| = 1} \langle \matvec{w}^*, \matvec{v} \rangle$, $\Pi$ is the set of possible learner policies, and $\textnormal{diam} \, \matvec{\mu}(\Pi) = \sup_{\pi_1, \pi_2 \in \Pi} \| \matvec{\mu}(\pi_1) - \matvec{\mu}(\pi_2)\|$.

    Furthermore, if $\textnormal{rank}(\bar{\matvec{A}}) = d$, then identifying a policy $\pi$ which matches the feature expectations in all perspectives exactly ensures that the learning agent matches the expert's performance, i.e., $$| \langle \matvec{w}^*, \matvec{\mu}(\pi^\expert) - \matvec{\mu}(\pi^\learner) \rangle| = 0.$$
\end{theorem}

The above theorem highlights that learning in the active third-person IL setting via feature-matching in the available perspectives can be successful if the perspectives provide sufficient information regarding feature-matching in the ground truth features.
Intuitively, performance degrades with decreasing information about the ground truth features retained in the perspectives.
Importantly, inaccuracies in feature-matching can have an amplified impact on the learning agent's performance as characterized by $1/\sigma(\bar{A})$.

\textbf{Non-linear reward functions or non-linear transformations.}
The characteristics of the problem change significantly for non-linear reward functions or non-linear transformations.
\begin{theorem}
    \label{thm:non-linear-rewards}
    Assume a non-linear reward function, i.e., $r(s,a)=g(\feats)$, where $g\colon \mathbb{R}^d \rightarrow \mathbb{R}$ is non-linear.
    Then, there exists an instance of the active third-person IL problem in which the relative decrease of the learner's performance is unbounded even if the learner matches the feature trajectories perfectly in each perspective and the perspectives jointly contain all information about the ground truth feature occurrences.
    The same is true for non-linear transformations of the ground truth features even if the rewards are linear in the original features and the transformation is bijective.
\end{theorem}
The above two statements indicate that the learner's performance is strongly influenced by the perspectives and the reward structure.
Nevertheless, we can identify rich settings in which the learner's performance can match that of the expert.
\begin{observation}
   \label{thm:reward-structure}
   Consider reward functions of the form $r(s,a) = \sum_{i=1}^K \matvec{w}^*_i g_i(\feats_{S_i})$,
   where $\feats_{S_i}$ is the subset of ground truth features indicated by the set $S_i \subseteq [d]$, $g_i\colon \mathbb{R}^{|S_i|} \rightarrow \mathbb{R}$ are possibly non-linear functions of the subsets of features, and $\matvec{w}^* = [\matvec{w}^*_1, \ldots, \matvec{w}^*_K]^\transpose \in \mathbb R^K$ are unknown reward parameters. 
   If the learner can observe the expert from perspectives $\perspectives=\{ p_i\colon \mathbb{R}^d \rightarrow \mathbb{R}^{|S_i|}\}_{i=1}^K$, where $p_i\colon \feats \mapsto \matvec{A}_i \feats_{S_i}$ and $\matvec{A}_i \in \mathbb{R}^{|S_i| \times |S_i|}$ is invertible, then the learner can asymptotically achieve the expert's performance via feature-matching.
\end{observation}
Reward functions with a structure according to the above observation, allow for non-linear bijective functions on a subset of the ground truth features regarding the reward function if the respective set of features can be observed via an invertible linear transformation in a single perspective.
The condition on the perspectives ensures that the learner can observe all dependencies among features that are reward-relevant.
Thus by matching the probabilities of all possible feature trajectories, the learner can achieve the expert's performance.
Note, however, that matching only the feature expectations from all perspectives is not sufficient to ensure matching the expert's performance~\cite{kamalaruban2019interactive,wulfmeier2015maximum}.

%% file: 4_methodology.tex
\section{Our Approach}
\label{sec:approach}

We first analyze a stylized variant of the active third-person IL problem in Section~\ref{sec:approach:warmup} and then develop our approach to the problem leveraging insights from this analysis in Section~\ref{sec:approach-details}.

\subsection{Warm up: Informative Perspectives}\label{sec:approach:warmup}

For our approach, we take inspiration from analyzing a stylized variant of the active third-person IL problem with rewards linear in some unknown ground truth features and with perspectives corresponding to known linear transformations $\linearpersp$, cf.\ Section~\ref{sec:analysis}.
In particular, assume that if the learning agent selects perspective $\perspective_t$ for the next demonstration, it observes the cumulative features $\bm o_t = \bm A_{\perspective_t} \cumfeats^\expert + \bm \eta_t$, where $\cumfeats^\expert = \mathbb{E}[\sum_{t=0}^{H-1} \bm \phi(s_t,a_t) | \pi^\expert]$ are the expected cumulative features of the expert policy $\pi^\expert$ and $\bm \eta_t$ is a subgaussian random variable (representing the possible randomness in a single demonstration and any other observational noise).
An approach to solving the third-person IL problem is then to select perspectives that allow accurate estimation of $\cumfeats^\expert$ as quickly as possible.
Concretely, if we would employ penalized least-squares regression to estimate $\cumfeats^\expert$, i.e., 
$
    \hat{\cumfeats}^E_t = \arg \min_{\cumfeats \in \mathbb{R}^d} \sum_{t'=1}^t (\bm o_{t'} - \bm A_{\perspective_{t'}} \cumfeats)^2 + \lambda \| \cumfeats\| _2^2,
$
where $\lambda > 0$ is a regularization factor, it can be guaranteed with a high probability that $\|\hat{\cumfeats}^\expert_t - \cumfeats^\expert\|_{\bm V_t}^2 = (\hat{\cumfeats}^\expert_t - \cumfeats^\expert)^\transpose \bm{V}_t (\hat{\cumfeats}^\expert_t - \cumfeats^\expert) \leq \beta_t$, where $\bm V_0 = \lambda \mathbf I$, $\bm V_t=\lambda \mathbf I + \sum_{t'=1}^t \bm A_{\nu_{t'}}^\transpose \bm A_{\nu_{t'}}$ and $\beta_t \geq 1$ is an increasing sequence of constants~\cite{lattimore2020bandit}.
If $\beta_t$ does not grow too quickly, then $\|\hat{\cumfeats}^\expert_t - \cumfeats^\expert\|_{\bm V_t}^2$ will shrink.
Furthermore, if perspectives are selected such that the smallest eigenvalue of $\bm V_t$ increases fast enough, $\|\hat{\cumfeats}^\expert_t - \cumfeats^\expert\|_2^2$ will also decrease.
This, in turn, implies that matching the feature expectations in a perspective to the respective observed empirical feature expectations will result in cumulative rewards close to those of the expert.

The volume of possible $\cumfeats^\expert$ satisfying $\|\hat{\cumfeats}^\expert_t - \cumfeats^\expert\|_{\bm V_t}^2 \leq \beta_t$ is proportional to $\beta_t / \prod_i \sigma_i(\bm V_t)$, where $\sigma_i(\bm V_t)$ is the $i$\textsuperscript{th} eigenvalue of $\bm V_t$.
Hence, in the absence of knowledge about the reward structure, for a fixed sequence $\beta_t$, a sensible perspective selection strategy could aim to maximize $\prod_i \sigma_i(\bm V_t)$ or $\log \det \bm V_t$---see~\cite[Chapter 21]{lattimore2020bandit} for a connection to D-optimal design.
Importantly, selecting strongly similar perspectives might result in a shrinkage of the volume of the ellipse containing the possible $\cumfeats$ (confidence ellipse) only along specific directions. 
Hence, it is crucial to account for the similarity of perspectives in the perspective selection strategy.
However, in most realistic settings, we will not know the relations of the different perspectives, i.e., $\linearpersp$ is unknown.
In such cases, we can still leverage insights from above: The size of the confidence ellipse depends on the selected perspectives and in particular, the relation of the transformations.
Thus, we should select perspectives containing complementary information as measured by the matrices $\bm A_{\perspective_t}$.
As we do not know these matrices, we suggest accounting for similarities/correlations among perspectives through prior knowledge and appropriate designs of the neural network architectures, thereby enabling more effective querying of complementary perspectives. We refer the reader to Section~\ref{sec:approach-details} for details and the experiments in Section~\ref{sec:experiments} for the advantages of doing so.

\subsection{Learning Algorithm}
\label{sec:approach-details}
\textbf{Overview.}
We aim to find suitable perspective selection strategies $\strategy$ and an approach allowing for effective learning from multiple perspectives. To this end, we propose an approach building on generative adversarial imitation learning (GAIL)~\cite{ho2016generative}, recent work on third-person IL~\cite{stadie2017third, torabi_generative_2019}, and our insights from Section~\ref{sec:approach:warmup}.

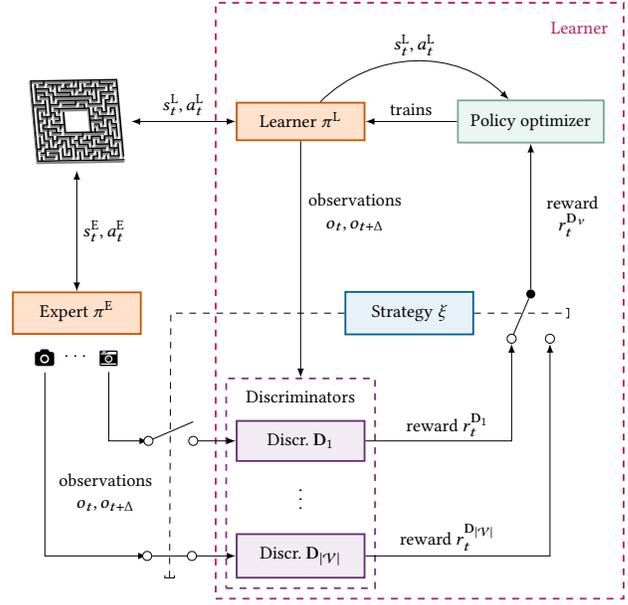
\begin{figure}%
\centering
\scalebox{0.85}{\input{tikz/framework2.tex}}

\captionof{figure}{Proposed approach for active third-person IL. 
For every perspective, the discriminator measures the imitation success for this perspective (here, visualized with one discriminator for each perspective).
The perspective selection strategy $\strategy$ is used to decide which perspective $\perspective$ to use for further training.}
\label{fig:framework}
\end{figure}

A schematic illustration of our approach is presented in Figure~\ref{fig:framework} and contains two novel elements: 
\begin{enumerate*}[label=(\roman*)]
    \item discriminators $\discriminator_1, \ldots, \discriminator_{|\perspectives|}$, one for each perspective $\perspective \in \perspectives$, and
    \item a perspective selection strategy~$\strategy$.
\end{enumerate*}
The discriminators measure how well the learner's policy $\pi^\learner$ matches the expert's policy $\pi^\expert$ in the respective perspective.
The different discriminators are not necessarily distinct neural networks but other suitable discriminator architectures can be used (see details below).
The perspective selection strategy $\strategy$ selects a perspective $\perspective$ from which expert and learner data is generated. After the discriminator has been trained for a fixed number of episodes on the expert's and learner's trajectories $\omega^E_\perspective, \omega^L_\perspective$ in perspective $\perspective$, respectively, the learner collects trajectories for updating its policy $\pi^\learner$. Before each observation, the learner must use the perspective selection strategy $\strategy$ to decide on a perspective $\perspective$. Algorithm~\ref{alg:algo_short} summarizes the interplay between the components.

Below, we provide detailed information about the components of our approach.

\textbf{Discriminator Architectures.} \label{sec:approach:discriminators}
Central to our approach is using discriminators $\discriminator_\perspective$ that can distinguish between expert and learner data for all available perspectives $\perspective \in \perspectives$. We present four choices for defining such discriminators within our framework. All discriminators are based on the DCGAN architecture~\cite{radford2015unsupervised} combined with design choices from \citet{stadie2017third} and \citet{torabi_generative_2019}\footnote{Using the architectures from \citet{stadie2017third} and \citet{torabi_generative_2019} without modification did not yield satisfactory results in our setting.}. In particular, we substitute missing action information by considering two slightly time-shifted observations $\bm o_t, \bm o_{t+\Delta}$, where $\Delta$ is the time shift, as inputs to a discriminator. We use binary cross entropy as the objective function for all discriminators. All layers are regularized with spectral normalization~\cite{miyato2018spectral} to improve stability and training performance. 
We consider the following possible designs for the discriminators:
\begin{enumerate*}[label=(\roman*), font=\itshape, leftmargin=15pt]
  \item Using a single discriminator network for all perspectives without conditioning on perspective information; 
  \item using an individual discriminator network for each perspective;
  \item using discriminators that combine perspective information with the features of the convolutional encoder of the network (this could be, e.g., the index of a perspective or its rotation parameters); and
  \item using a conditional discriminator based on the FiLM architecture~\cite{perezFiLMVisualReasoning2017} which scales the outputs of its convolutional layers based on perspective-specific weights.    
\end{enumerate*}
Motivated by the analysis in Section~\ref{sec:approach:warmup}, we account for correlations between perspectives with a parameter-shared correlation network when using multiple discriminators. The correlation network receives the current discriminator's CNN features as input and conditions on the current perspective $\perspective$ through a one-hot encoding of $\perspective$.

\textbf{Perspective selection.}\label{desc:perspective_selection}
Observing the expert only from a single perspective may limit the learner's performance as this perspective might not be informative about the expert's behavior (see also Section~\ref{sec:approach:warmup}). To enable learning from multiple perspectives, we assume that in each iteration of the algorithm, the learner can choose a perspective $\perspective \in \perspectives$ from which it observes the expert. A discriminator $\discriminator_\perspective$ is trained to differentiate between expert and learner for perspective $\perspective$. The discriminator's output is then used as a reward signal for the learner. The selection between the available perspectives is accomplished by the perspective selection strategy $\strategy$. The selection strategy $\strategy$ is employed in two components of our algorithm:
\begin{enumerate*}[label=(\roman*)]
    \item to select a perspective $\perspective \in \perspectives$ from which expert and learner data is generated for discriminator optimization and
    \item to select a perspective, i.e., a discriminator $\discriminator_\perspective$ used to provide rewards for the learner's training.
\end{enumerate*}
Note that we assume that the learner can select an individual perspective through the perspective selection strategy $\strategy$ for each of its internal training episodes. While numerous selection strategies are conceivable, in this work, we consider uniform random,  feature correlation-based, and UCB-like strategies (cf.\ Section~\ref{sec:experiments}).

\textbf{Learner policy optimization.}
The output of the discriminator $\discriminator_\perspective$, i.e., a measure of how likely the expert generated the data, is used as a reward signal for training the  learner's policy $\pi^\learner$. For each trajectory the learner generates, the perspective selection strategy $\strategy$ determines the discriminator providing the reward. We use PPO~\cite{schulman2017ppo} as a policy optimization algorithm due to its good performance and stability on a wide range of tasks.

\hfill

\input{algorithms/algorithm.tex}

%% file: tikz/framework2.tex
\tikzstyle{policy}=[draw=colororange!80!black, thick, fill=colororange!20!white, minimum width=2.cm, inner sep=2mm]
\tikzstyle{optimizer}=[policy, draw=colorturquise!80!black, fill=colorturquise!20!white]
\tikzstyle{strategy}=[policy, draw=colormyblue!80!black, fill=colormyblue!10!white]
\tikzstyle{discriminator}=[policy, draw=colorpurple!80!black, fill=colorpurple!10!white]
\tikzstyle{switch}=[circle,fill=none,draw=black,inner sep=0.5mm, fill=white]

\begin{tikzpicture}
  \small 
  \node[minimum height=1.5cm, draw=none] (env) {\includegraphics[width=1.5cm]{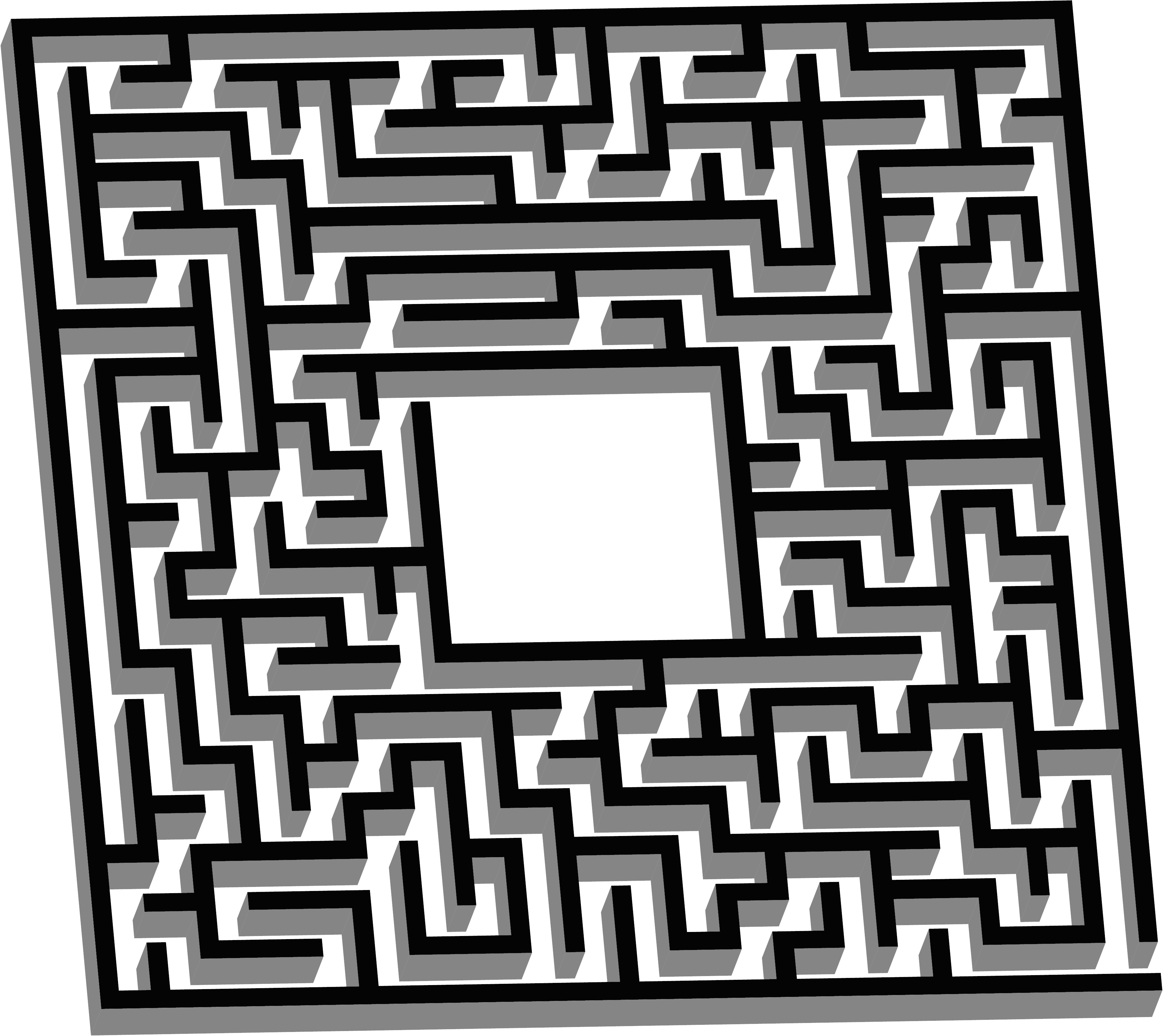}};

  \node[policy, right of=env, xshift=2.5cm] (learner) {Learner \smash{$\pi^\learner$}};

  \draw[<->, >=latex] (learner) -- node[pos=0.5, above] () {$s_t^\learner, a_t^\learner$} (env);

  \node[optimizer, right of=learner, xshift=2.6cm] (optimizer) {Policy optimizer};
  \draw[->, >=latex] (optimizer) -- node[pos=0.5, above] () {trains} (learner);
  \draw[->, >=latex] (learner) to [out=45,in=135] node[pos=0.5, above] () {$s_t^\learner, a_t^\learner$} (optimizer);
  
  \node[policy, below of=env, yshift=-2cm] (expert) {Expert \smash{$\pi^\expert$}};
  \draw[<->, >=latex] (expert) -- node[pos=0.5, right] () {$s_t^\expert, a_t^\expert$} (env);

  \node[strategy, right of=expert, xshift=42mm] (strategy) {Strategy $\strategy$};
  
  \node[discriminator, below of=learner, yshift=-4cm] (discriminator1) {Discr.\ $\discriminator_1$};
  \node[below of=discriminator1, yshift=2mm] (discriminator2) {$\vdots$};
  \node[discriminator, below of=discriminator2, yshift=-0cm] (discriminator3) {Discr.\ $\discriminator_{|\perspectives|}$};
  
  \node[draw=colorpurple, dashed, thick, inner sep=1.5mm, label={[label distance=-5mm]above:Discriminators}, fit={([shift={(0mm,5mm)}]discriminator1.north) (discriminator2) (discriminator3)}] (discriminators) {};

  \node[circle, below of=optimizer, fill=black, inner sep=0.5mm, yshift=-17mm] (switch) {};
  \draw[draw=black] (switch) -- node[pos=1.0,switch] (switch-r1) {} ($(switch) + (-3mm,-7mm)$);
  \draw[draw=none] (switch) -- node[pos=1.0,switch] (switch-r2) {} ($(switch) + (+3mm,-7mm)$);
  \node[draw=none, right of = discriminator1, yshift = 0.25cm,xshift=1.3cm] (label1) {reward $r_t^{\discriminator_1} $};
\node[draw=white, right of = discriminator3, yshift = 0.3cm,xshift=1.3cm] (label1) {reward $r_t^{\discriminator_{|\perspectives|}}$};

  \draw[->, >=latex, draw=black] (discriminator1.east) to  ($(discriminator1 -| switch-r1)$) to ($(switch-r1.south)$);
  
  \draw[draw=black,dashed,-Bracket] (strategy.east) -- ($(strategy.east -| switch-r2) + (3mm,0mm)$);

  \draw[->, >=latex, draw=black] (discriminator3.east) to  ($(discriminator3 -| switch-r2)$)
  to ($(switch-r2.south)$) ;

  \draw[->, >=latex] (switch) --node[pos=0.5, right] (rewardbox) {\parbox{1.1cm}{\centering reward $r_t^{\discriminator_\perspective}$}}  (optimizer);
  
  \draw[->, >=latex] (learner) -- node[pos=0.3, right] () {\parbox{1.5cm}{\centering observations $o_t,o_{t+\Delta}$}} (discriminators);
  
  \node[draw=colorpink!80!black, dashed, thick, inner sep=1.5mm, label={[yshift=-6mm,xshift=25mm]above:\textcolor{colorpink!80!black}{Learner}}, fit={([shift={(0mm,14mm)}]learner.north west) (rewardbox) (optimizer) (discriminators) }] (learner-block) {};
  
  \node[below of=expert, yshift=3mm, xshift=-5mm] (perspective1) {\faCamera};
  \node[below of=expert, yshift=3mm, xshift=5mm] (perspective2) {\faCameraRetro};
  \node[below of=expert, yshift=3mm] () () {$\cdots$};
  
  \draw[draw=black, ->, >=latex] (perspective1.south) -- (perspective1 |- discriminator3) --++(+16mm,0mm) node[pos=1.,switch] (break1) {};
  \coordinate (foo) at ($(break1) + (3.5mm,0mm)$) ;
  \draw[->, >=latex] (break1) --++(+7mm,0mm)  node[pos=1.,switch] () {}-- (discriminator3);
  
  \draw[draw=black, ->, >=latex] (perspective2.south) -- (perspective2 |- discriminator1) --++(6.2mm,0mm) node[pos=1.,switch] (break1) {};
  \draw[->, >=latex] ($(break1) +(+7mm,0mm)$) -- node[pos=0.,switch] () {} (discriminator1);
  \draw[draw=black] (break1) -- ($(break1) + (7mm,3mm)$);
  
  \coordinate (ce) at ($(break1) + (3.5mm,0mm)$) ;
  \draw[draw=black,dashed,-Bracket] (strategy.west) -- ($(strategy.west -| ce)$) -- ($(foo) + (0mm,-3.5mm)$);

  \node[above of=foo, xshift=-10mm] () {\parbox{2.2cm}{\centering observations\\$o_t,o_{t+\Delta}$}};

\end{tikzpicture}

%% file: algorithms/algorithm.tex
\begin{algorithm2e}
\caption{Active third-person IL}\label{alg:algo_short}
\SetAlgoLined
\DontPrintSemicolon
\KwIn{perspective selection strategy $\strategy$, nr.\ of iterations $K$, perspectives $\perspectives$}
\tcc{Initialisation}
Initialize discriminators $\discriminator_\perspective$, environment $\environment$, expert policy $\pi^\expert$, learner policy $\pi^\learner$, and perspective selection strategy $\strategy$\;
\tcc{learning}
\For{$i=1,\ldots,K$}{
   $\perspective_i \leftarrow$ Next perspective according to persp.\ selection strategy $\strategy$ \;
   $\omega^\expert_{\perspective_i} \leftarrow$ Demonstration(s) from expert's policy $\pi^\expert$ \;
   $\omega^\learner_{\perspective_i} \leftarrow$
    Trajectories from learner's policies $\pi^\learner$ \;
    Update discriminator $\discriminator_{\perspective_i}$ based on $\omega^\expert_{\perspective_i}, \omega^\learner_{\perspective_i}$\;
    Update learner's policy by training in the environment using rewards  $-\log(\discriminator_{\perspective})$ where $\perspective$ is selected according to $\strategy$ \;
}%
\KwResult{Optimized learner's policy $\pi^\learner$} %
\end{algorithm2e}

%% file: 6_experiments.tex
\section{Experiments}
\label{sec:experiments}

In this section, we empirically evaluate our proposed approach. 
We first demonstrate the feasibility of third-person imitation learning in a simplified setting on tabular environments in Section~\ref{sec:feasibility}. 
Then, we introduce the considered environments and tested perspective selection strategies in Sections~\ref{sec:experiments:environments} and ~\ref{sec:experiments:selection_strategies}, respectively, followed by our conducted main experiments in Section~\ref{sec:experiments:evaluation}. Additional results and details, e.g., on environments and hyperparameters, optimizers, and network architectures, are provided in the Appendix.

\subsection{The Tabular Case With Known Dynamics}
\label{sec:feasibility}

Here, we demonstrate that actively selecting the perspectives from which to observe the expert can lead to improved performance in comparison to selecting perspectives uniformly at random in grid world environments.
Experimenting with grid worlds has the advantage that we can perform imitation learning using linear programming, avoiding many of the challenges that we face when using our GAIL-based approach.
In particular, we consider grid worlds of size $10 \times 10$ in which an agent can collect different types of objects, each having some randomly chosen reward in $[0,1]$.
In total, we consider 4 types of objects each of which is present in the grid world 2 times.
The ground-truth features on which the rewards are defined are indicators of whether a cell contains a particular type of object.
Upon collection of a reward object, the object is replaced and the agent is randomly placed at an empty position in the grid world.
For the perspectives, we consider subsets of the indicator features or random linear transformations of the features.
To amplify the effect of active selection strategies, we ensure the presence of highly similar perspectives---in the case of indicator features, we duplicate the feature vector for the first object 12 times, and in the case of random linear transformations, we create a total of 40 random transformations making it probable that some perspectives are similar. 
Such similar (but to some degree \emph{redundant}) perspectives would be selected by a uniform strategy with a fixed probability while an adaptive active learning strategy should avoid them.
Feature matching is performed using the linear program presented in the Appendix assuming full knowledge about the environment dynamics. %
Error terms for the individual perspectives are scaled proportionally to how often the individual perspectives have been selected.
We consider 3 active learning strategies based on different levels of knowledge about the perspectives (\textsc{active (corr)}, \textsc{active (sim)}, \textsc{active (var)}; details are provided in the Appendix) and compare them to a uniform selection strategy (\textsc{uniform}). %
We evaluate the different strategies regarding the achieved cumulative ground-truth reward.
Additional details regarding the experimental setup are provided in the Appendix. %

\begin{figure}[!t]
  \subcaptionbox{Basis-vector transformations}[.5\linewidth]{%
    \includegraphics[width=\linewidth,scale=0.5]{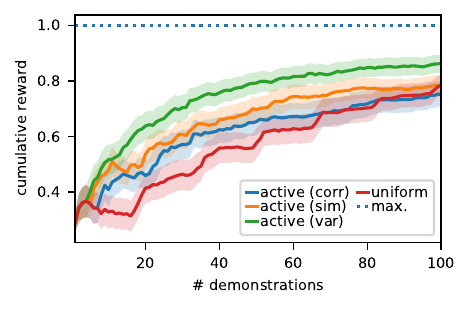}%
  }%
  \hfill
  \subcaptionbox{Random linear transform}[.5\linewidth]{%
    \includegraphics[width=\linewidth,scale=0.5]{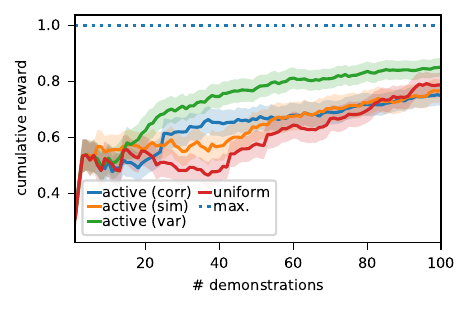}%
  }
   \vspace{-2mm}
    \caption{Performance of active learning strategies on grid worlds in comparison to naive uniform perspective selection. The active learning strategies clearly outperform the baseline. See the main text for details.}
  \label{fig:gw-basic}
  
\end{figure}

Our results showing the performance of the different perspective selection strategies regarding the attained reward over the number of observed perspectives are presented in Figure~\ref{fig:gw-basic}.
Results are averaged over 100 random grid worlds and cumulative rewards are normalized so that the maximum achievable reward is $1$. 
The shaded regions indicate 95 \% confidence intervals regarding the mean reward.
The active learning strategies clearly outperform the naive uniform perspective selection strategy for the first $\sfrac{2}{3}$ of observed demonstrations, highlighting the utility of collecting and leveraging information from different perspectives in a principled way.
For more observed demonstrations, \textsc{uniform} catches up with \textsc{active (corr)} and \textsc{active (sim)}.
The strategy \textsc{active (var)} dominates all other strategies for all number of observed demonstrations.

\subsection{Benchmark Environments}\label{sec:experiments:environments}
For our further experiments, we evaluate our approach on 2 environments (Point and Reacher) described below.
To evaluate whether effective perspective selection occurs we include non-informative perspectives that always return a constant observation. Other perspectives contain partial information and require implicitly combining multiple perspectives over multiple interactions, e.g., projections on the x- and y-axis to imitate the expert. Performance is again measured using the ground truth reward. 

\begin{figure}[!htb]
  \begin{subfigure}[t]{0.40\columnwidth}
    \centering
    \includegraphics[width=.65\textwidth]{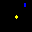}
    \caption{Point: Baseline}
    \label{fig:point:all1}
  \end{subfigure}%
  \begin{subfigure}[t]{0.40\columnwidth}
    \centering
    \includegraphics[width=.65\textwidth]{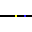}
    \caption{Point: x-perspective}
    \label{fig:point:x1}
  \end{subfigure}\\[1mm]
  \hfill
  
  \begin{subfigure}[t]{0.40\columnwidth}
    \centering
    \includegraphics[width=.65\textwidth]{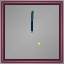}
    \caption{Reacher: Baseline}
    \label{fig:reacher:all1}
  \end{subfigure}%
  \begin{subfigure}[t]{0.40\columnwidth}
    \centering
    \includegraphics[width=.65\textwidth]{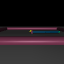}
    \caption{Reacher: side persp.}
    \label{fig:reacher:x1}
  \end{subfigure}%
  \vspace{-2mm}
  \caption{Examples of different perspectives available in the considered environments. \emph{(\subref{fig:point:all1},\subref{fig:reacher:all1})} Perspectives including all information available in the environment; \emph{(\subref{fig:point:x1},\subref{fig:reacher:x1})} Perspectives showing only partial information (e.g., 
  the projection onto the x-axis as in \emph{(\subref{fig:point:all1})}).}
  \label{fig:env_perspectives}
\end{figure}

\textbf{Point.}
The agent controls a point mass in a plane and aims to move it towards a target location, cf.\ Figure~\ref{fig:point:all1}. The following $4$ perspectives are available to the learner:
\begin{enumerate*}[label=(\roman*),font=\itshape]
  \item \emph{birds-eye perspective} in the form of a 2D image containing all information (Figure~\ref{fig:point:all1});
  \item / \item \emph{x-perspective} and \emph{y-perspective} in the form of 1D vectors corresponding to a horizontal and a vertical projection of the birds-eye perspective, respectively (see Figure~\ref{fig:point:x1} for the x-perspective);
  \item \emph{no-information perspective} corresponding to a black image.
\end{enumerate*}
We use a rule-based expert policy that walks perfectly toward the goal.

\textbf{Reacher.} In the \emph{MuJoCo reacher environment}~\citep{brockman2016openai}, a two-jointed robot arm aims to point its end towards a (randomly spawning) target on the plane, see Figure~\ref{fig:reacher:all1}. 
The learner can select the following perspectives:
\begin{enumerate*}[label=(\roman*),font=\itshape]
    \item \emph{birds-eye perspective} in the form of a 2D image containing all information (Figure~\ref{fig:reacher:all1});
  \item / \item \emph{side-perspectives} in the form of 2D images looking at the reacher arm from the side (see Figure~\ref{fig:reacher:x1} for an example). The difference between both perspectives is a 90° rotation;
  \item \emph{no-information perspective} based on an angle where the robot arm is not visible.  
\end{enumerate*}
As an expert policy, we use an MLP policy trained to maximize the ground truth reward with PPO~\cite{schulman2017ppo}.

\subsection{Perspective selection strategies}\label{sec:experiments:selection_strategies}

In each algorithm iteration, the perspective selection strategy $\strategy$ selects a perspective used to observe expert and learner trajectories for discriminator optimization, cf.\ Algorithm~\ref{alg:algo_short}. During learner training, the perspective is chosen anew for each episode. Refer to Section~\ref{sec:approach-details} for a detailed description. We evaluate the following $3$ selection strategies:

\begin{enumerate}[label=(\roman*), font=\itshape, leftmargin=15pt]
  \item \textbf{Uniform strategy (\textsc{Uniform}).} Uniform random sampling among perspectives.

\item \textbf{UCB style strategy (\textsc{UCB}).} This strategy is inspired by the upper confidence bound-based algorithms commonly used in the bandit literature~\cite{lattimore2020bandit}. In the $t$\textsuperscript{th} trajectory the strategy $\strategy$ selects the discriminator as 
$
    \discriminator_t \in \argmin_{\discriminator_v , v \in \perspectives} \; \discriminator_v(\Tilde{\omega}^\learner) - c \sqrt{\sfrac{\log(t)}{N_{\perspective,t}}},
$
where $N_{\perspective,t}$ is the number of times perspective $\perspective$ has been used so far, $c$ is a hyper-parameter, and $\discriminator_v(\Tilde{\omega}^\learner)$ is the probability that the observations in a batch of observations stem from the expert. The strategy focuses on improving the imitation performance on perspectives leading to large discrimination errors, accounting for the number of times the perspectives have been considered so far.

\item \textbf{Feature dissimilarity strategy (\textsc{Dissimilarity}).} A strategy that selects perspectives based on the similarities of the perspective's features. It tracks the approximate discounted feature expectations for each perspective and uses an exponentially weighted average to account for the non-stationarity of the policy during training. A perspective $\perspective_t$ at step $t$ is sampled with a probability proportional to the inverse correlation coefficient between $\perspective_t$ and all other perspectives. A more detailed description is provided in the Appendix. %
\end{enumerate}

\subsection{Discriminators}
\label{sec:experiments:conditional_discriminators}

As the discriminator determines the reward for the RL agent, it is of key importance in our framework. To understand how different design choices for the discriminators impact the performance of our approach, we experiment with four different architectures. Details for the conditional discriminator architectures can be found in the Appendix. %
Concretely, we consider the following architectures in line with Section~\ref{sec:approach}:
\begin{enumerate}[label=(\roman*), font=\itshape, leftmargin=15pt]
  \item \textbf{Multiple discriminators (\textsc{Multiple}).} One discriminator for each perspective.

  \item \textbf{Single discriminator (\textsc{Single}).} A single discriminator for all perspectives without any additional information.

  \item \textbf{Conditional discriminator (\textsc{Conditional}).} A single discriminator for all perspectives that conditions on the current perspective by concatenating perspective information with features extracted from the discriminator's convolutional layers. For Reacher, we use the camera's angle and distance as information. For Point, we use the index of the selected perspective.
  
  \item \textbf{FiLM discriminator (\textsc{FiLM}).} The FiLM architecture \cite{perezFiLMVisualReasoning2017} allows conditioning a network on arbitrary information by generating conditional weights that are used to scale a convolutional network's feature maps. In our case, this is perspective information such as, e.g., a perspective's rotation angles. 
\end{enumerate}

\subsection{Empirical Evaluation}

\label{sec:experiments:evaluation}
We use PPO~\cite{schulman2017ppo} to optimize the learner and tune relevant hyperparameters with a grid search for each environment (details are provided in the Appendix). All experiments use the same 20 seeds. For evaluation, we use the rliable library \cite{agarwalDeepReinforcementLearning} to plot the interquartile mean with bootstrapped confidence intervals computed from $50{,}000$ subsampled runs. On Reacher, learning is unstable, with failing runs generating very large negative values. We therefore cap the negative reward at $-300$ to prevent outliers from distorting our results. We present our findings arranged in three scenarios that pose increasingly challenging problems to the agent. In the \emph{easy} scenario, the agent is presented with the partially informative perspectives described in Section~\ref{sec:experiments:environments} and a single non-informative perspective. The \emph{duplicate} scenario is the same as \emph{easy}, with each perspective duplicated thrice. Lastly, in the \emph{adversarial} scenario, we present the agent with a set of uninformative perspectives among which a single fully informative perspective (Reacher) or two partially informative perspectives (Point) must be chosen.

\begin{figure}[!t]
  \subcaptionbox{Strategies\label{fig:point_easy_a}}[.48\linewidth]{%
    \includegraphics[width=42mm]{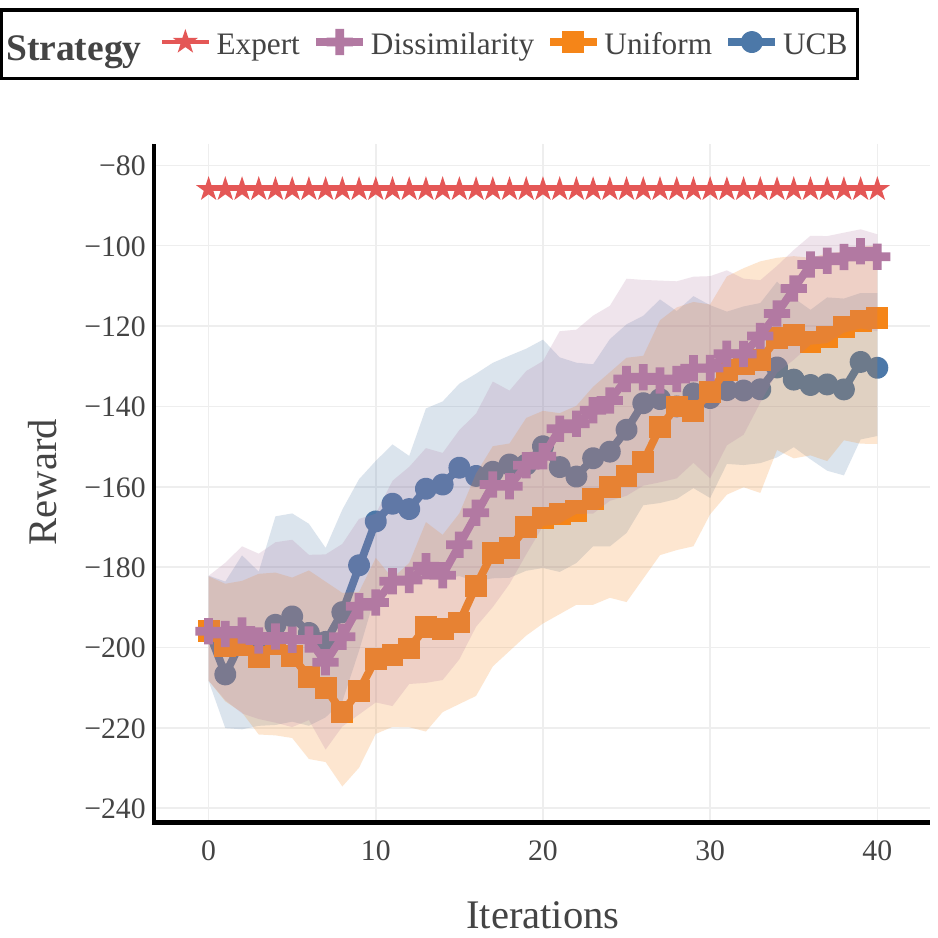}%
  }%
  \hfill
  \subcaptionbox{Discriminator architectures\label{fig:point_easy_b}}[.48\linewidth]{%
    \includegraphics[width=42mm]{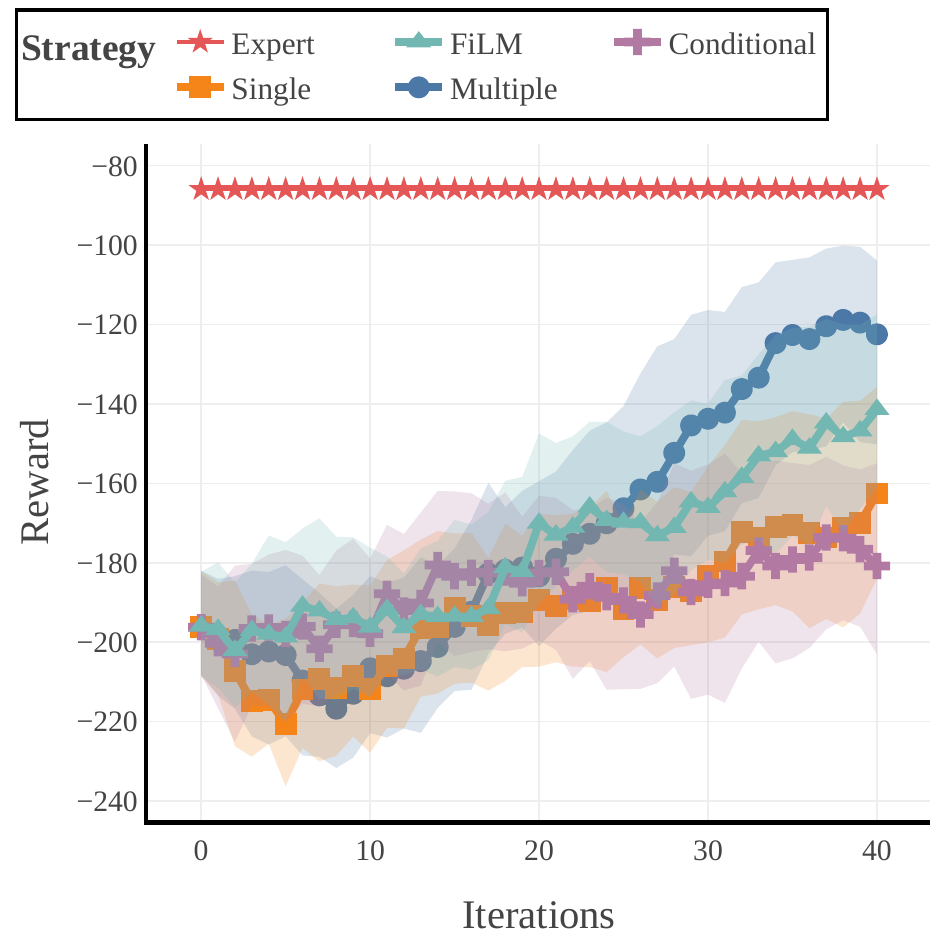}%
  }
    \caption{Comparison of perspective selection strategies and discriminator architectures for Point in the \emph{easy} scenario. \emph{a)} All strategies effectively learn when using \textsc{multiple} discriminators. The feature \textsc{Dissimilarity} perspective selection strategy performs best. \emph{b)} Using a single discriminator for each perspective (\textsc{multiple}) yields the highest reward, followed by the \textsc{FiLM} network. All runs use the \textsc{Uniform} strategy.}
  \label{fig:point_easy}
  
\end{figure}

\textbf{Selection strategies and discriminator architectures in the \emph{easy} scenario.}\label{par:point_results_easy}
We evaluate all perspective selection strategies $\strategy$ and discriminator architectures introduced in Section~\ref{sec:experiments:selection_strategies} and~\ref{sec:experiments:conditional_discriminators} on Reacher and Point. The expert's reward serves as an upper bound on performance.  Effective perspective selection using a particular strategy $\strategy$ is evaluated by introducing a non-informative perspective offering no information to the agent. See Appendix~\ref{app:environment-details} for details on these perspectives. 
Figure~\ref{fig:point_easy} presents our findings for the Point environment. We first ablate the effect of the selection strategy when using a unique discriminator for each perspective (Figure~\ref{fig:point_easy_a}). All strategies effectively learn from multiple perspectives. The perspective selection based on feature \textsc{Dissimilarity} performs best. 
We select perspectives uniformly at random with fixed seeds to evaluate the benefits of the different considered discriminator architectures without confounding effects from the perspective selection strategies (Figure~\ref{fig:point_easy_b}).
Having a single discriminator for each available perspective yields the highest reward, followed by using the \textsc{FiLM} architecture.
Using a \textsc{Single} discriminator without perspective information or with naive conditioning (\textsc{Conditional}) on the perspective performs worst.
For Reacher, we defer the reader to Appendix~\ref{app:additional_results:reacher_reasy_results}. The results for Reacher are not as clear as for Point. All strategies exhibit some learning, with the \textsc{Dissimilarity} strategy performing worst. Using \textsc{multiple} discriminators or the \textsc{FiLM} network achieves similar performance.

\begin{figure}[!t]
  \subcaptionbox{Point}[.49\linewidth]{%
    \includegraphics[width=1.05\linewidth]{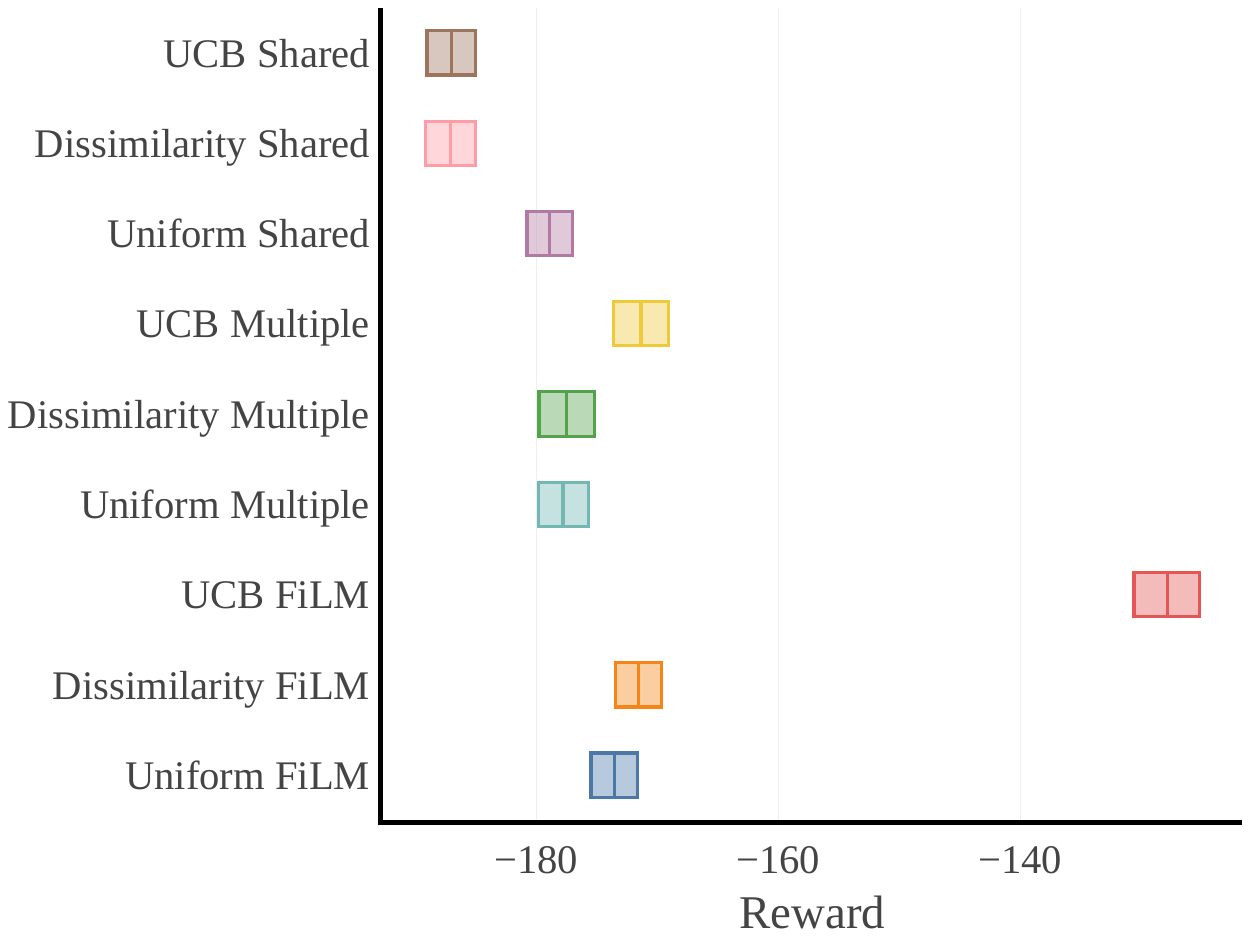}%
  }%
  \hfill
  \subcaptionbox{Reacher}[.49\linewidth]{%
    \includegraphics[width=1.05\linewidth]{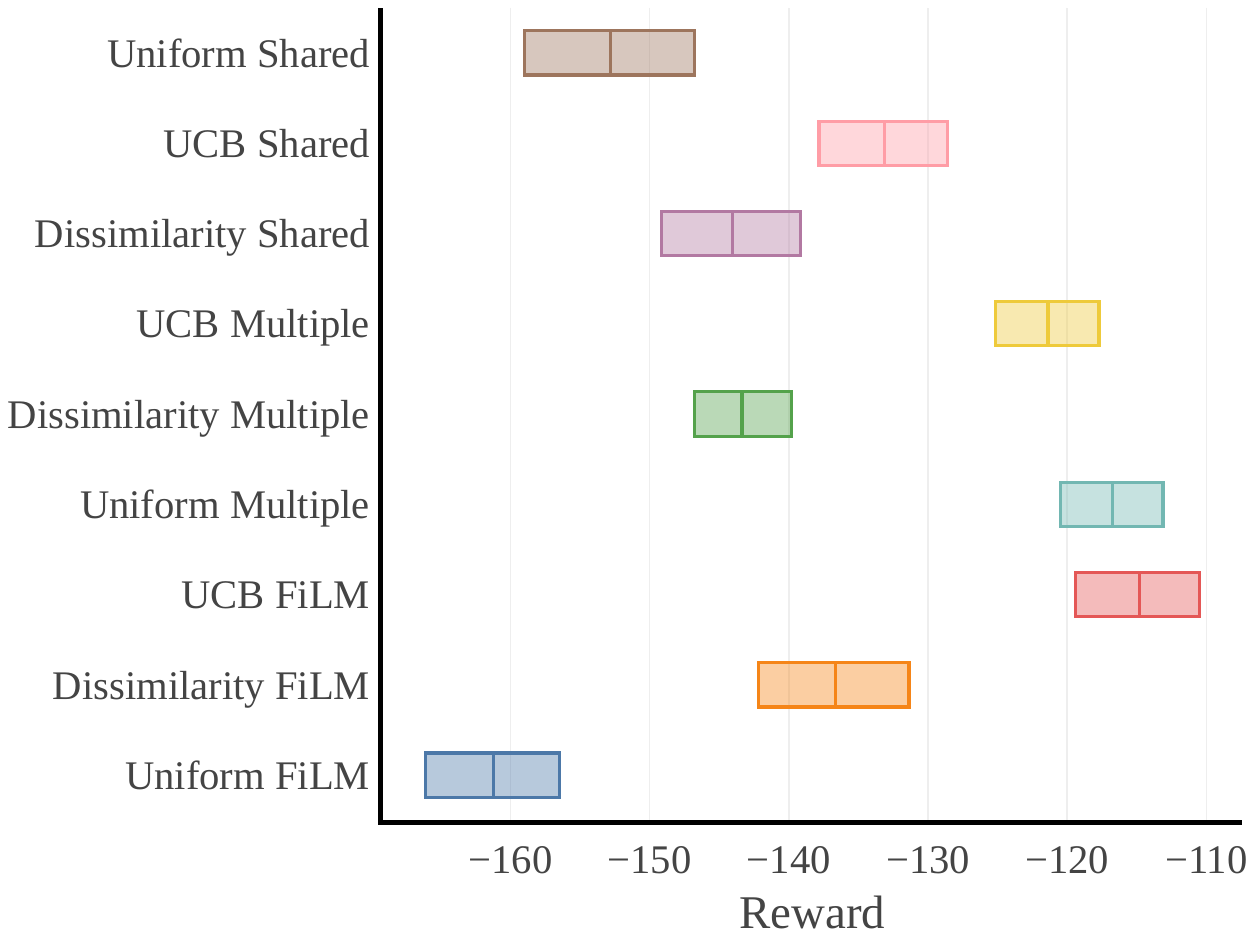}%
  }
    \caption{Results for strategy and discriminator combinations in the duplicate environment configuration. \textsc{UCB FiLM} is the best combination for both environments with a large margin to the second best agent on Point.}
  \label{fig:duplicate_perspectives}
\end{figure}

\textbf{Results for the \emph{duplicate} scenario.}\label{par:results_duplicate}
Section~\ref{sec:approach:warmup} motivates the exploitation of similarities/correlations between perspectives to accelerate imitation learning. 
Information about such structure is not necessarily available or present in the learning environment: For Point introduced in Section~\ref{sec:experiments:environments}, no such correlations exist in the perspectives. %
However, we can easily induce relations among perspectives by duplicating perspectives. In this scenario, we duplicate all of the partially informative perspectives 3 times to test how well strategies and discriminator architectures perform in the presence of correlated perspectives. Figure~\ref{fig:duplicate_perspectives} depicts our findings.
Surprisingly, the parameter-sharing heuristic (\textsc{Shared}) we use to learn correlations (see Section~\ref{sec:approach:discriminators}) does not yield improvements over using multiple isolated discriminators without parameter-sharing. 
Combining the \textsc{UCB} strategy with the \textsc{FiLM} discriminator results in the highest reward for both environments. On Point, there is a large gap between all other methods, whereas on Reacher \textsc{Uniform} and \textsc{UCB} with \textsc{Multiple} discriminators provide similar results.

\textbf{Results for the \emph{adversarial} scenario.}\label{par:results_adversarial}
Part of learning successfully in the active third-person IL problem boils down to effectively reducing the number of times an uninformative perspective is selected. We verify whether different combinations of selection strategies and discriminator architectures can do this through a challenging setup where most perspectives in the environment are uninformative. To this end, we present the agent with a fully informative perspective and 5 uninformative perspectives in Reacher. For Point, the agent can choose from the x-axis and y-axis perspectives and 6 uninformative perspectives.
Analyzing the results, no single best strategy works well for Reacher and Point. We hypothesize that this is due to both environments' different properties posing distinct challenges to a learning algorithm. For Reacher, the \textsc{Dissimilarity Multiple} combination works best, whereas \textsc{Uniform} strategies fail to learn. This highlights the importance of active perspective selection in this environment. \textsc{UCB} selection combined with the \textsc{FiLM} discriminator performs strongly on Point, whereas it fails when it is combined with \textsc{Multiple} discriminators. Note that there is a substantial drop in overall performance compared to the \emph{easy} and \emph{duplicate} scenarios for both environments, particularly for Point.

\begin{figure}[!t]
  \subcaptionbox{Point}[.49\linewidth]{%
    \includegraphics[width=\linewidth]{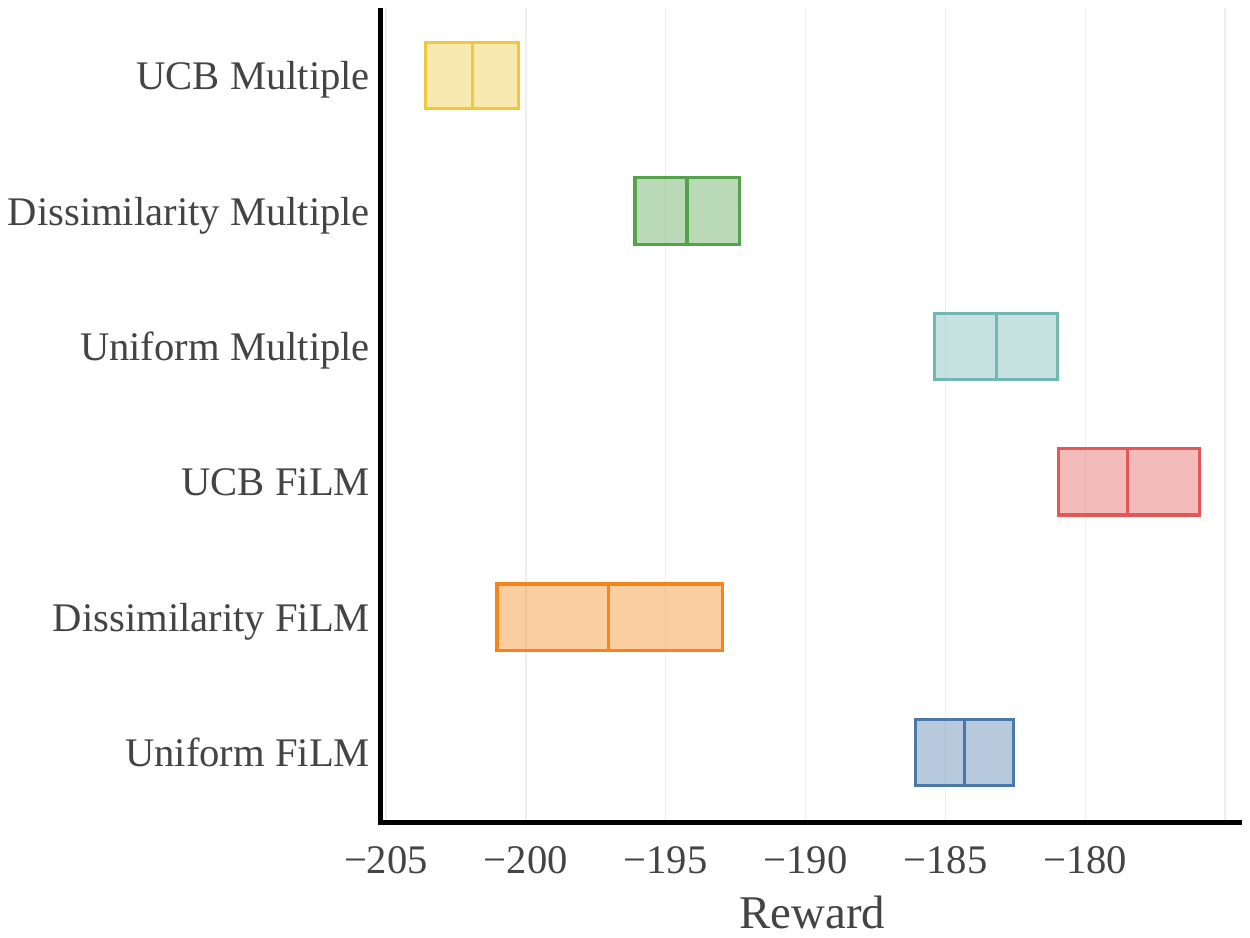}%
  }%
  \hfill
  \subcaptionbox{Reacher}[.49\linewidth]{%
    \includegraphics[width=\linewidth]{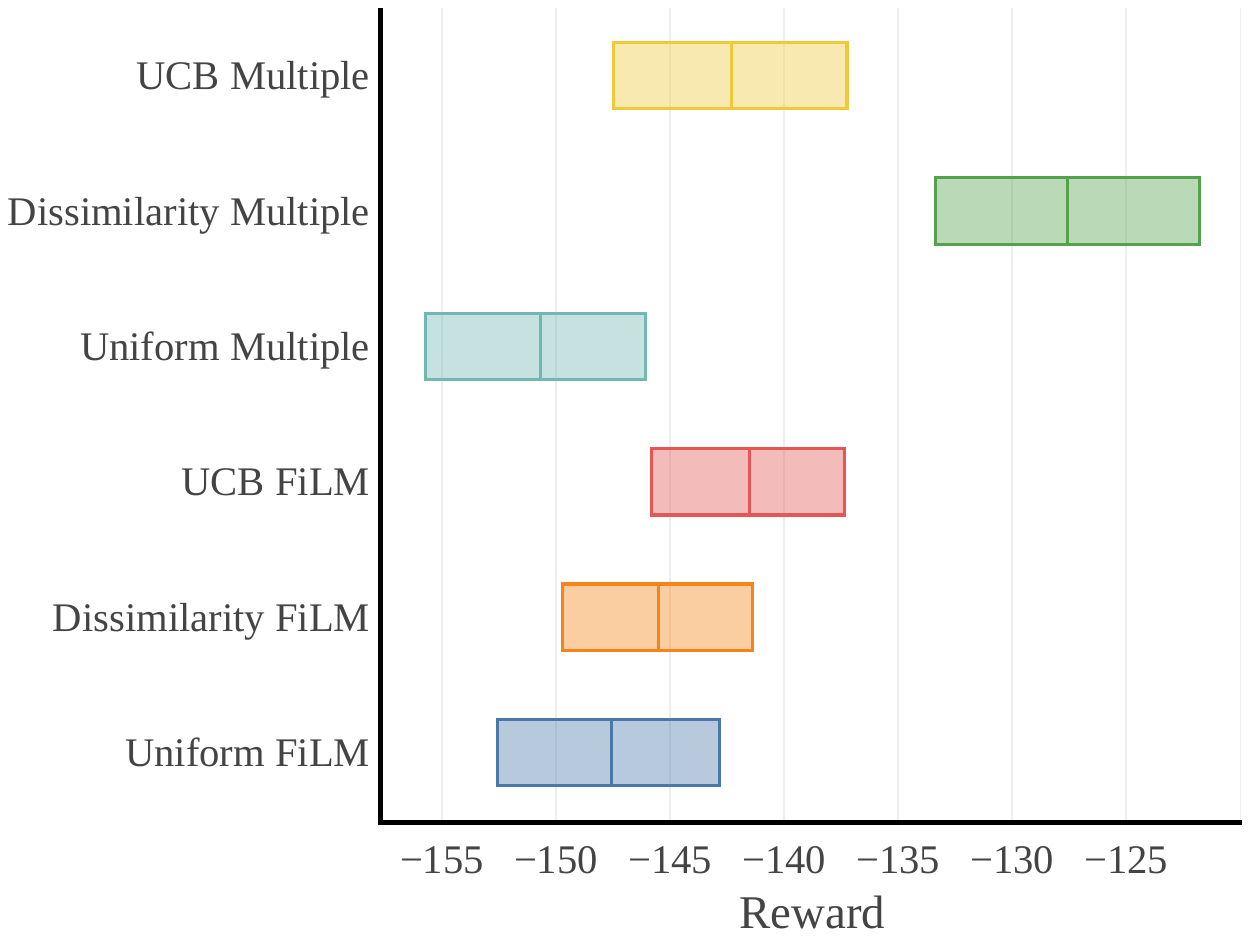}%
  }
    \caption{Results for strategy and discriminator combinations in the adversarial environment configuration. The best combination depends on the environment. \textsc{UCB FiLM} shows strong overall performance. For Reacher, using \textsc{Dissimilarity} sampling with \textsc{Multiple} discriminators performs best.}
  \label{fig:adversarial_perspectives}
\end{figure}

\textbf{Discussion.}
Considering the results described in the previous paragraphs, it first stands out that the \textsc{UCB} strategy combined with the \textsc{FiLM} discriminator performs either best or second best in all scenarios and environments. Successfully imitating the expert in the active third-person IL setting amounts to a more frequent selection of informative perspectives while avoiding uninformative ones. A naturally following hypothesis from our findings would be that \textsc{UCB FiLM} is able to avoid uninformative perspectives. %
As shown in Appendix~\ref{app:additional_results:point_selection_frequencies} this is indeed the case.
We further note that our proposed framework interleaves three highly noisy, non-stationary processes: Perspective selection, agent training, and discriminator training. We hypothesize that the intricate and often implicit interplay between all these components is quickly destabilized, rendering stable and robust learning highly challenging. Additionally, the similarity of the provided perspectives plays a role. For Reacher, the fully informative perspective is very dissimilar from the non-informative perspectives with a correlation coefficient between images of around $0.06$ (see Figure~\ref{fig:envreacher} for a visualization). Inspecting Figure~\ref{fig:envpoint} on the other hand, we can see that for Point, the non-informative perspective is largely the same as the informative perspectives. This fact may explain the success of the \textsc{Dissimilarity} strategy for Reacher in the adversarial scenario.

%% file: 7_conclusion.tex
\section{Conclusions and Future Work}
\label{sec:conclusions}

We introduced the active third-person imitation learning problem, a challenging variant of the learning from demonstrations problem, in which the learning agent has control over the perspective from which it observes an expert's demonstration. We formalized this problem and analyzed its characteristics. In particular, we showed that for linear reward functions, learning via feature matching is feasible, provided perspectives are given by linear transformations. Additionally, we found that feature matching does not guarantee successful learning in the case of non-linear reward functions. Inspired by generative adversarial imitation learning, we proposed an approach for solving the active third-person IL problem. We evaluated our method in toy and benchmark environments on increasingly difficult scenarios. Our findings indicate that various perspective selection strategies and discriminator architectures enable learning from perspectives with partial information. 

While our approach to the active third-person IL problem makes a step towards learning from demonstrations when different perspectives are available, we have only considered scenarios where the set of perspectives is finite. An exciting direction for future work is the generalization of active third-person IL to unbounded sets of perspectives, e.g., allowing the learner to freely choose the camera angle from which it observes an expert. Another interesting problem setting emerges when the learner can change the perspective while receiving a single demonstration. Lastly, we want to explore more efficient methods for parameterizing the ensemble of discriminators when the number of perspectives $|\perspectives|$ is large, e.g., through Hypernetworks \cite{haHyperNetworks2016}.

%% file: a8_appendix-proofs.tex
\section{Proofs}

\subsection{Proof of Theorem~\ref{thm:linear-rewards-and-projections}}

\begin{proof}
    The proof of the first part of the statement follows by observing that $\| \linearpersp( \bm \mu(\pi^\expert) - \bm \mu(\pi^\learner))\| < \epsilon / |\perspectives|$ implies that $\|\bar{\bm A}( \bm \mu(\pi^\expert) - \bm \mu(\pi^\learner))\| < \epsilon$.
    The result then follows by invoking Theorem 1 from~\cite{haug2018teaching}.

    The second part of the statement follows by observing that if $\epsilon=0$, $\text{rank}(\bar{A}) = d$, we have $\sigma(\bar{A}) > 0$ and $\rho(\bar{A}, w^*) = 0$.
\end{proof}

\subsection{Proof of Theorem~\ref{thm:non-linear-rewards}}

\begin{proof}
      We start by providing an example of a non-linear reward function (conjunction of two features) for which expert performance cannot be achieved in the third-person IL setting.
      To this end, consider an MDP with action set $\actions=\{\text{left}, \text{right}\}$, 2-dimensional features, a horizon $H=2$, and perspectives corresponding to observing only a single of these features.
      The reward function is
      \begin{align*}
          r(s) = \begin{cases}
            1 & \textnormal{if } \bm \phi(s) = [1,1]^T, \textnormal{and}\\
            0 & \text{otherwise}.
          \end{cases}
      \end{align*}
      The dynamics and features of the MDP are shown in Figure~\ref{fig:mdp-performance-bound}.
      \begin{figure}[H]
          \centering
          \input{tikz/mdp-performance-bound.tex}
          \caption{MDP}
          \label{fig:mdp-performance-bound}
      \end{figure}
      The agent is assumed to always start in state $S_0$ at the beginning of an episode.
      An optimal policy $\pi^*$ would always perform action "left", resulting in an expected cumulative reward of $0.5$.

      Observe that the probabilities of all possible feature trajectories in each perspective are identical for any possible policy, i.e.,
      \begin{align*}
        p(\bm \phi(S_{t=0})_1 = 2, \bm \phi(S_{t=1})_1 = 0 | \pi) &= 0.5 \pi(a_1=\text{left}) + 0.5 \pi(a_1=\text{right})  \\
          &= 0.5 [\pi(a_1=\text{left}) + \pi(a_1=\text{right})] \\
          &= 0.5, \\
        p(\bm \phi(S_{t=0})_1 = 2, \bm \phi(S_{t=1})_1 = 1 | \pi) &= 0.5, \\
        p(\bm \phi(S_{t=0})_2 = 2, \bm \phi(S_{t=1})_2 = 0 | \pi) &= 0.5, \\
        p(\bm \phi(S_{t=0})_2 = 2, \bm \phi(S_{t=1})_2 = 1 | \pi) &= 0.5, 
      \end{align*}
      where $S_{t=t'}$ is the random variable representing state $S$ at time $t'$, and $\bm \phi(S_{t=t'})_d$ denotes the features associated with state $S_{t=t'}$ in dimension $d$.
      Hence the observations in a single perspective carry no information whatsoever to distinguish between different policies and how well they match the expert's features while perfectly matching the feature expectation marginals (even matching the actual sample distribution).
      Hence a learning agent with a policy that would in the first time step take action "right", would exactly match the feature trajectory distributions in each dimension individually but achieve a cumulative reward of $0$.

      Now, assuming that the reward function is linear in some ground truth features $\bm \phi'(s,a)$.
      Assume a 5-state MDP in which only a single state provides a reward of $1$.
      The feature function is linear in features corresponding to a one-hot encoding of the states.
      For these features we can construct a bijective mapping to the feature vectors of the MDP shown above. 
      Assuming the same dynamics and reward assignment as in the above MDP concludes the proof.
\end{proof}

%% file: tikz/mdp-performance-bound.tex
\tikzstyle{state}=[circle, draw=gray, minimum width=1cm]

\begin{tikzpicture}
  \node[state] (S_0) {$S_0$};
  \node[below of=S_0, yshift=2mm] () {$[2,2]^T$};

  \node[state, left of=S_0, xshift=-10mm, yshift=10mm] (S_1) {$S_1$};
  \node[above of=S_1, yshift=-2mm] () {$[1,1]^T$};

  \node[state, left of=S_0, xshift=-10mm, yshift=-10mm] (S_2) {$S_2$};
  \node[below of=S_2, yshift=2mm] () {$[0,0]^T$};

  \node[state, right of=S_0, xshift=10mm, yshift=10mm] (S_3) {$S_3$};
  \node[above of=S_3, yshift=-2mm] () {$[0,1]^T$};

  \node[state, right of=S_0, xshift=10mm, yshift=-10mm] (S_4) {$S_4$};
  \node[below of=S_4, yshift=2mm] () {$[1,0]^T$};

  \node[left of=S_0, anchor=east] () {"left"};
  \node[right of=S_0, anchor=west] () {"right"};

  \draw[->,>=latex] (S_0) to[out=135,in=0] node[pos=0.5, yshift=3.5mm] () {$50 \%$} (S_1);
  \draw[->,>=latex] (S_0) to[out=-135,in=0] node[pos=0.5, yshift=-3.5mm] () {$50 \%$} (S_2);

  \draw[->,>=latex] (S_0) to[out=45,in=-180] node[pos=0.5, yshift=3.5mm] () {$50 \%$} (S_3);
  \draw[->,>=latex] (S_0) to[out=-45,in=-180] node[pos=0.5, yshift=-3.5mm] () {$50 \%$} (S_4);

\end{tikzpicture}

%% file: a9_additional_results.tex
\section{Perspective selection frequencies on Point}
\label{app:additional_results:point_selection_frequencies}
\begin{figure*}[!t]
  \subcaptionbox{Point}[.45\linewidth]{%
    \includegraphics[width=\linewidth,scale=0.5]{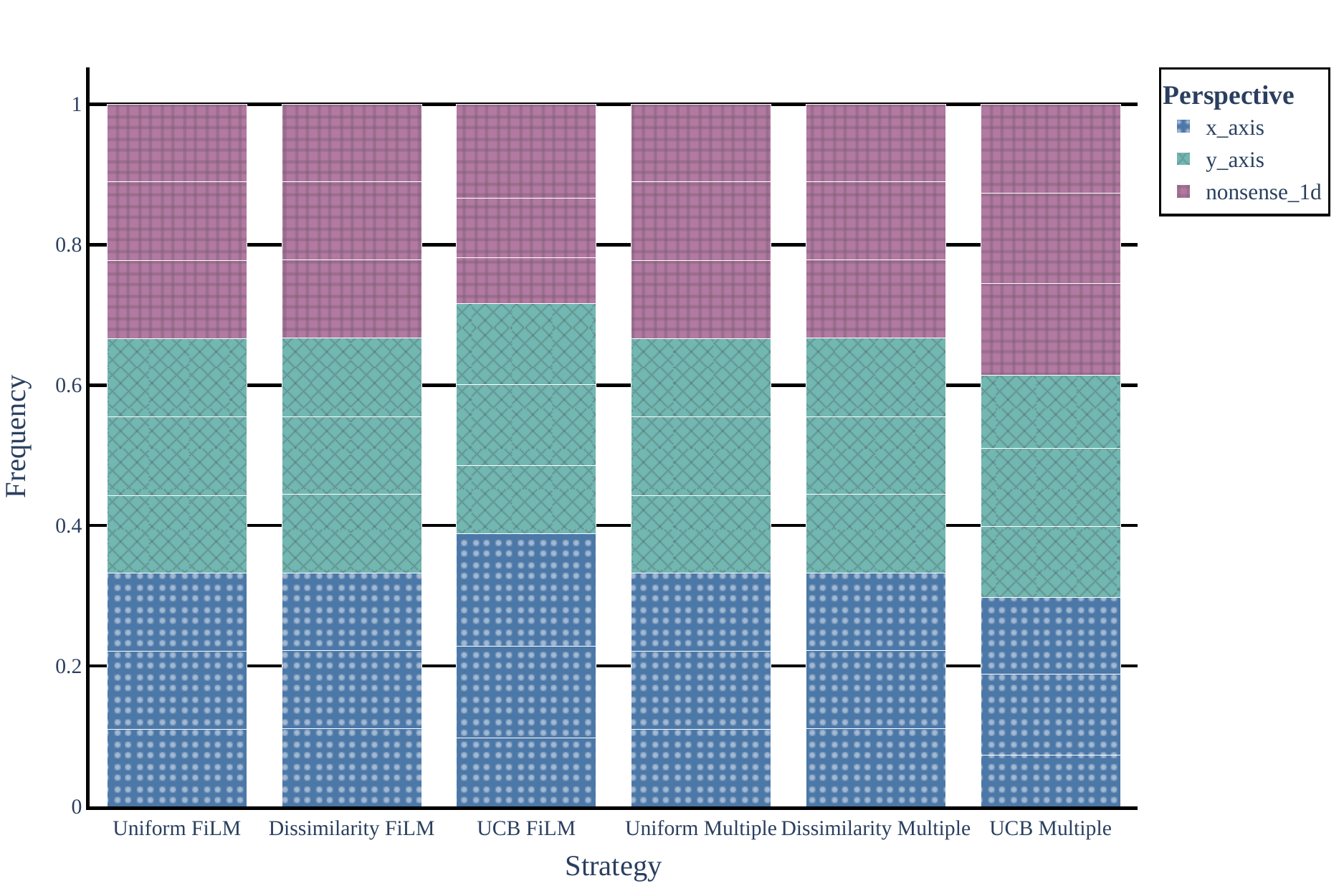}%
  }%
  \hfill
  \subcaptionbox{Reacher}[.45\linewidth]{%
    \includegraphics[width=\linewidth,scale=0.5]{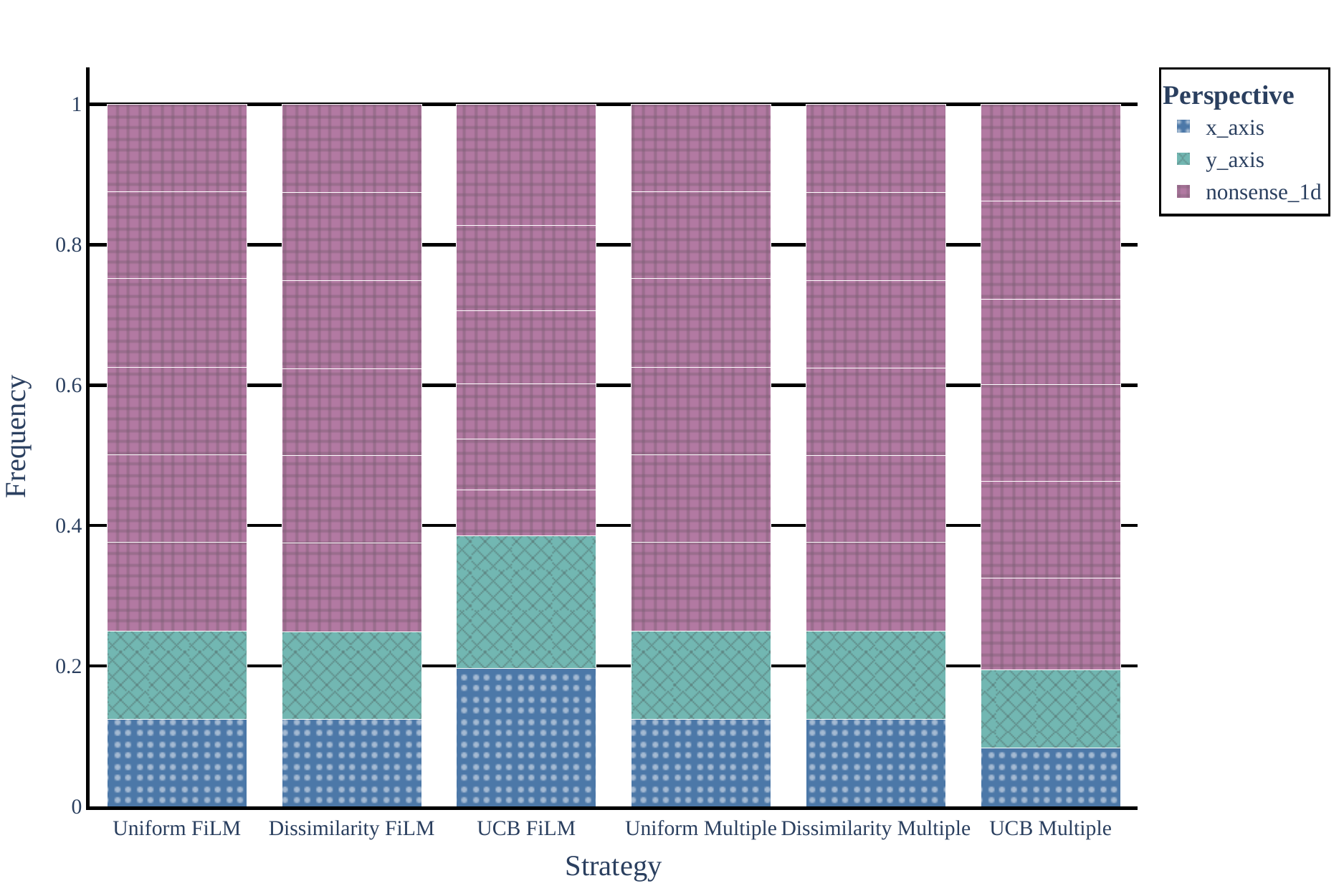}%
  }
    \caption{Results for strategy and discriminator combinations in the adversarial environment configuration. Which combination performs best is highly dependent on the environment.}
  \label{fig:point:selection_freqs}
\end{figure*}

This section briefly studies the hypothesis that successful imitation in the active third-person IL setting amounts to avoiding uninformative perspectives.
Figure~\ref{fig:point:selection_freqs} shows the selection frequencies for Point in the \emph{duplicate} and \emph{adversarial} scenarios. Indeed, we find that the best combination of discriminator and strategy (\textsc{UCB FiLM}) selects the uninformative perspective less often than other configurations. In the \emph{duplicate} scenario, the \textsc{Uniform} probability of selecting an uninformative perspective is $33\%$. Compared to this, the \textsc{UCB FiLM} combination only selects the nonsense perspective with  $28.4\%$ probability, which is an improvement of $14.7\%$.
For the \emph{adversarial} scenario, the improvement is even more stark: Here, the probability of selecting an uninformative perspective with \textsc{Uniform} random selection is 6 out of 8 or $75\%$. In contrast, the selection probability for \textsc{UCB FiLM} is only $61.4\%$, yielding an $18.3\%$ improvement. However, the results also show that the selection frequencies alone cannot explain agent performance. While \textsc{UCB FiLM} reduces the chance of sampling an uninformative perspective in both the \emph{duplicate} and \emph{adversarial} scenarios, imitation performance does not increase proportionally. Confounding factors at the intersection between discriminator training, agent training, and perspective selection are likely.

\section{Learning curves for Reacher in the \emph{easy} scenario}
\label{app:additional_results:reacher_reasy_results}
\begin{figure}[!t]
  \subcaptionbox{Strategies}[.5\linewidth]{%
    \includegraphics[width=\linewidth]{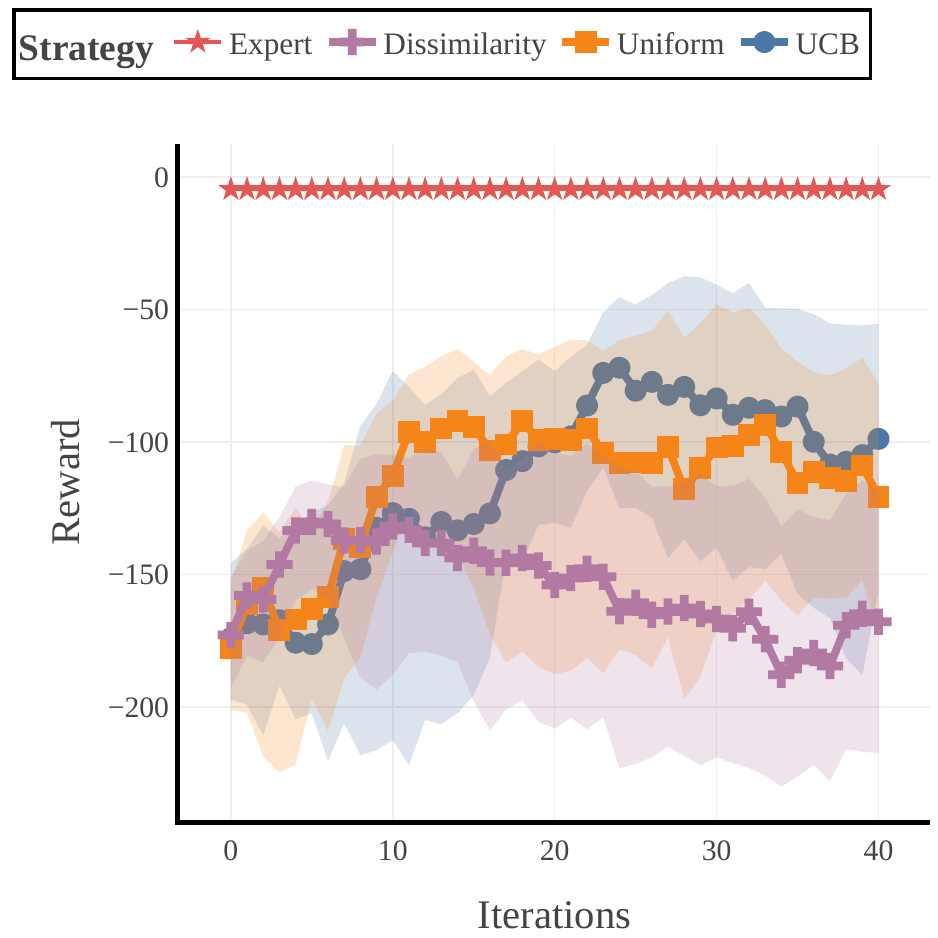}%
  }%
  \hfill
  \subcaptionbox{Discriminator architecture}[.5\linewidth]{%
    \includegraphics[width=\linewidth]{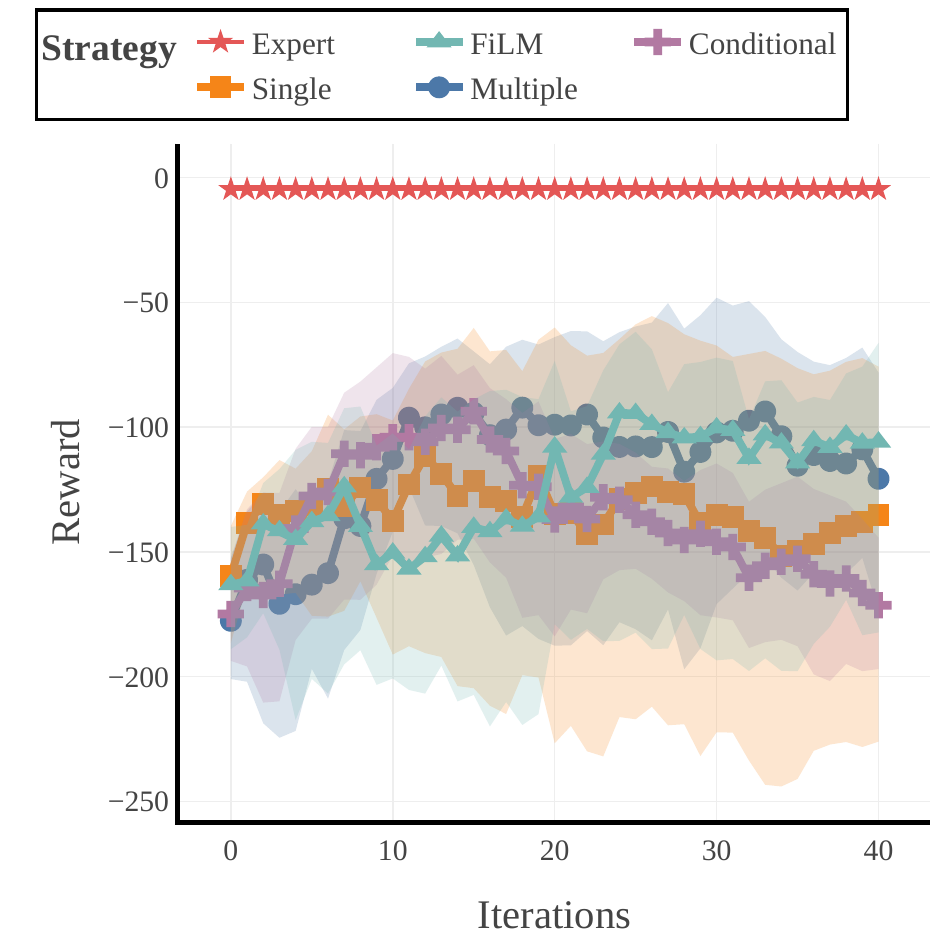}%
  }
    \caption{Using \textsc{Multiple} discriminators or conditioning on the \textsc{FiLM} architecture \cite{perezFiLMVisualReasoning2017} performs best in both environments.}
  \label{fig:reacher_easy}
\end{figure}

This section presents learning curves for perspective selection strategies and discriminator architectures on Reacher. The results are not as clear-cut as on Point and suffer a high variance. We find that this is due to discriminator training being less stable.
Evaluating strategies in Figure~\ref{fig:reacher_easy}, the \textsc{UCB} and \textsc{Uniform} strategies perform roughly equally well, with the \textsc{Dissimilarity} strategy not improving over random performance. This finding is striking as it starkly contrasts the results presented in Paragraph~\ref{par:results_adversarial}. We hypothesize that it is a consequence of the uninformative perspective in Reacher having the most dissimilar features out of the three used perspectives in the \emph{easy} scenario. In the \emph{adversarial} scenario, the fully informative perspective is the most dissimilar from a set of identical uninformative perspectives.
We see a similar picture for the discriminators to Point: \textsc{Multiple} and \textsc{FiLM} perform best. The \textsc{Single} discriminator is competitive with the best settings at the cost of a very high variance. Naively conditioning on perspective information (\textsc{Conditional}) does not work on Reacher, just as it does not work on Point.

%% file: a10_env_descriptions.tex
\section{Detailed Environment Descriptions}\label{app:environment-details}
We use 2 environments for our experiments. Details of common hyperparameters, such as the maximum episode length or the observation space, can be found in Section~\ref{app:hyperparams_network_compute:hyperparamaters} and Table~\ref{tab:parameters:point} and~\ref{tab:parameters:reacher}.  For each environment, we provide 4 perspectives: A baseline perspective showing all information, two perspectives providing partial information, and a perspective with no relevant information. As a performance metric for our evaluation, we use the reward provided by the environment.

\subsection{Point Environment} \label{app:point:details}
Point is a 2-dimensional environment where the agent must move a yellow point (the \emph{chaser}) to a blue point (the \emph{goal}) within a $[-5,5] \times [-5,5]$ coordinate system (the \emph{arena}). The chaser always starts an episode at the origin $[0,0]$. The goal is created at a random position within an Euclidean distance of $4.5$ to the origin. Figure~\ref{fig:envpoint} visualizes the observation space, and Figure~\ref{fig:point:all} the arena in particular. The entire environment specification is:
\begin{itemize}
    \item \textbf{Action space}: Point has a 2D continuous action space for movements in the $x$ and $y$ direction. In particular, movements in both directions are possible and bounded by a magnitude of $0.1$ per dimension and per time step, i.e., $(a_1, a_2) \in [-0.1,0.1]^2$.
    \item \textbf{State space}: As states, we provide the agent with the current position of the chaser, the distance between chaser and goal $d_\text{cg}$ \footnote{The largest distance possible is reached when goal and chaser reside on diagonally opposite corners.}, and the position of the goal in this episode in the coordinate system. Concretely $(x_\text{chaser},y_\text{chaser},d_\text{cg},x_\text{goal},y_\text{goal}) \in [-5, -5, 0, -5, -5] \times [5, 5, 10\sqrt{2}, 5, 5]$.
    \item \textbf{Environment's reward function}: The reward function is defined using the negative of the Euclidean distance between the current position and the goal.
    \begin{gather}\label{eq:point_env_reward}
        r(s) = - \Big\lVert [x_\text{chaser}, y_\text{chaser}] - [x_\text{goal}, y_\text{goal}] \Big\rVert_2.
    \end{gather}
    \item \textbf{Observation space}: As the environment is only two-dimensional, partially informative perspectives are represented as 1D images. Using $C \times H \times W$ to specify the number of channels and size of an RGB image, we use $3 \times 32 \times 32$ images for the fully informative perspective and $3 \times 32$ images for the partially informative perspectives, respectively. The chaser has color $(255,255,0)$, corresponding to yellow, whereas the goal is blue, $(0,0,255)$, and the background is black, $(0,0,0)$.
\end{itemize}

\subsubsection{Perspectives}\label{app:point:details:perspectives}
\begin{figure}[ht]
  \centering
  \begin{subfigure}[t]{0.2\textwidth}
  \includegraphics[width=1.\textwidth]{figures/envs_images/point_org.png}
      \caption{Full information: 2D view}
    \label{fig:point:all}
  \end{subfigure}%
  \hspace{0.5cm}
  \begin{subfigure}[t]{0.2\textwidth}
  \includegraphics[width=1.\textwidth]{figures/envs_images/point_xaxis.png}
\caption{Partial information: $x$-Perspective}
\label{fig:point:x}
  \end{subfigure}%
  \hspace{0.5cm}
  \begin{subfigure}[t]{0.2\textwidth}
  \includegraphics[width=1.0\textwidth]{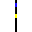}
        \caption{Partial information: $y$-Perspective}
    \label{fig:point:y}
  \end{subfigure}%
  \hspace{0.5cm}
  \begin{subfigure}[t]{0.2\textwidth}
  \includegraphics[width=1.0\textwidth]{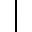}
        \caption{No information: Black image}
    \label{fig:point:nonsense}
  \end{subfigure}%
  \caption{Different perspectives for the Point environment. \textit{(\subref{fig:point:all}}) The baseline perspective includes all available information, \textit{(\subref{fig:point:x})} \& \textit{(\subref{fig:point:y})} show perspectives with partial information, and \textit{(\subref{fig:point:nonsense})} shows the no-information perspective, i.e., a black line.}
  \label{fig:envpoint}
\end{figure}

We define partial information perspectives through a projection\footnote{Details on how the network architecture is adapted are provided in Appendix~\ref{app:hyperparams_network_compute:network}.}
on the $x$ (resp. $y$) axis, cf.\ Figures~\ref{fig:point:x}, \ref{fig:point:y}. This entails both perspectives, including information about movement in one particular direction ($a_1, a_2$ respectively). We use a black 1D image as an uninformative perspective, allowing us to use this "nonsense" view of the environment in conjunction with the partially informative perspectives.

\subsubsection{Expert Policy}\label{app:point:details:expert_policy}
Since the observations include the current position and the position of the goal, we define an (almost) perfect deterministic expert policy\footnote{The expert is only almost perfect, as it does not stop moving when reaching the target. Instead, it oscillates around the target with small movements.} by taking a maximum step in the correct direction. Concretely:
\begin{align}\label{eq:point_expert}
    [a_1, a_2] = 0.1 \cdot \sign \big([x_\text{goal},y_\text{goal}] - [x_\text{chaser},y_\text{chaser}] \big)~.
\end{align}

\subsection{Reacher Environment}\label{app:reacher:details}
In Reacher~\cite{brockman2016openai}, a two-jointed robot arm tries to move its end, referred to as the \emph{fingertip}, by moving both joints toward a target on the 2-dimensional plane represented by a yellow point. This corresponds to the default configuration specified by \citet{brockman2016openai} and is visualized in Figure~\ref{fig:reacher:all}. In each episode, the goal spawns randomly within an Euclidean distance of $0.2$ to the origin.
We can define the environment as follows:
\begin{itemize}
    \item \textbf{Action space}: The reacher arm moves by applying torques to both hinge joints, concretely $\mathcal A \coloneqq [a_1, a_2] \in [-1, 1]^2$. This corresponds to a 2-dimensional continuous action space.
    \item \textbf{State space}: The state space contains the current state of the reacher arm as well as the absolute position of the target, the angular velocity of the arm, and the relative position of the reacher's fingertip to the target. Concretely, a state in Reacher is defined by an 11-dimensional vector containing: 
    \begin{itemize}
        \item $\sin$ and $\cos$ respectively of both parts of the arm (4)
        \item position of the target (2)
        \item angular velocity of both parts of the arm (2)
        \item 3D - vector between fingertip and goal in the form $(x,y,0)$ as they do never differ in the $z$-coordinate (3)
    \end{itemize}
    \item \textbf{Environment's reward function}: The reward consists of the sum of the distance between the fingertip of the robot arm and the goal with an added squared L2-penalty for taking too large actions. Concretely:
    \begin{gather}\label{eq:reacher_env_reward}
        r(s, a) = - \Big\lVert [x_\text{tip}, y_\text{tip}] - [x_\text{goal}, y_\text{goal}] \Big\rVert_2 - \sum_{i=1}^{|\mathcal A|}a_i^2~.
    \end{gather}
    \item \textbf{Observation space}: We experimented with $3 \times 32 \times 32$ ($C \times H \times W$) RGB image observations due to their appealing computational efficiency but observed almost no learning progress. Visual inspection of these small images revealed a lack of detail, leading us to use $3 \times 64 \times 64$ images. We also note that the side perspective displayed in Figure~\ref{fig:reacher:x} showed strong similarities between the originally defined goal color and the depicted purplish frame. Therefore, we decided to change the color of the goal to yellow.
\end{itemize}

\subsubsection{Perspectives}\label{app:reacher:details:perspectives}
\begin{figure}[ht]
  \centering
  \begin{subfigure}[t]{0.2\textwidth}
  \includegraphics[width=1.\textwidth]{figures/envs_images/reacher_org.png}
      \caption{Full information: Birds-eye view}
    \label{fig:reacher:all}
  \end{subfigure}%
  \hspace{0.5cm}
  \begin{subfigure}[t]{0.2\textwidth}
  \includegraphics[width=1.\textwidth]{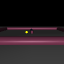}
\caption{Partial information: Front perspective}
\label{fig:reacher:x}
  \end{subfigure}%
  \hspace{0.5cm}
  \begin{subfigure}[t]{0.2\textwidth}
  \includegraphics[width=1.0\textwidth]{figures/envs_images/reacher_side.png}
        \caption{Partial information: Side perspective}
    \label{fig:reacher:y}
  \end{subfigure}%
  \hspace{0.5cm}
  \begin{subfigure}[t]{0.2\textwidth}
  \includegraphics[width=1.0\textwidth]{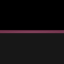}
        \caption{No information: Environment frame}
    \label{fig:reacher:nonsense}
  \end{subfigure}%
    \caption{Different perspectives for the Reacher environment. \textit{(\subref{fig:reacher:all})} birds-eye view including all relevant information, \textit{(\subref{fig:reacher:x})} \& \textit{(\subref{fig:reacher:y})} are perspectives with partial information, and \textit{(\subref{fig:reacher:nonsense})} shows a perspective with no information.}
  \label{fig:envreacher}
\end{figure}

Reacher is a 3D environment allowing us to control the amount of information shown by adapting the camera angle. As the reacher arm moves only 2-dimensionally on a plane, we define a baseline showing all information through a central birds-eye view of the environment, cf.\ Figure~\ref{fig:reacher:all}.  
Partially informative perspectives utilize a side view of the environment to obscure relevant information. We find that an angle of $7^\circ$ between the camera and the surface retains some relevant environment information while posing a sufficiently hard learning problem. Our two proposed partially informative perspectives show a view from the front and one side ($90^\circ$ rotation on the $z$-axis) using the $7^\circ$ angle between the surface and the camera, cf.\ Figures~\ref{fig:reacher:x} and~\ref{fig:reacher:y} for a visualization. As a non-informative perspective, we use an angle of $0^\circ$ between the camera and the reacher surface, thus showing only a side view of the environment's frame. 

\subsubsection{Expert Policy}\label{app:reacher:details:expert_policy}
As Reacher is more complex than Point, we cannot define a rule-based expert. Instead, we optimize our expert with PPO \cite{schulman2017ppo}. As a neural network, we use a 2-layer MLP with 32 hidden neurons and a Gaussian policy trained for 20000 epochs with batch size 1\,024. We verify the quality of the expert policy through visual inspection.

%% file: a11_correlation_selection.tex
\section{Dissimilarity Sampling}
\label{app:corr_diag_selection}

This section describes our perspective selection strategy based on estimating feature similarities between the different perspectives. It estimates the discounted feature expectations of a perspective-policy combination via sampling, i.e., it approximates
\begin{align*}
\matvec{\mu} (\perspective) = \mathbb{E} \Big[\sum_{t=0}^\infty \gamma^t \matvec{\phi}_\perspective (s_t,a_t) \mid \pi \Big]~, \perspective \in \perspectives,
\end{align*}
where $\pi$ is the policy for which the feature expectations are computed.
To account for the learner's policy $\pi^\learner$ being non-stationary, we use an exponentially weighted average to estimate the approximate discounted feature expectations. The exponential weights down-weigh earlier features from previous policies.
Our intuition behind formulating the \emph{Dissimilarity Sampling} (DS) strategy is that the agent should prefer to select perspectives that are as diverse as possible in order not to acquire \emph{redundant} information. The DS strategy achieves this by tracking approximate discounted feature expectations as detailed above and sampling from perspectives based on their inverse overall similarity to all other perspectives. Algorithm~\ref{alg:corr_diag_selection} outlines the procedure.
First, it ensures that each perspective has been selected at least once to ensure that valid similarities can be calculated (line 2). After an estimate $\hat{\matvec{\mu}}$ for the discounted feature expectations is available for each perspective, the DS strategy first calculates a score $s_i$ for each perspective $\perspective_i$ by inverting the summed correlation coefficients between $\perspective_i$ and all other perspectives (lines 5-6). These scores are then normalized to yield probabilities, from which the next perspective is sampled line (7). Including stochasticity in the selection process serves a dual purpose:
\begin{enumerate*}[label=(\roman*), font=\itshape]
    \item It prevents an early lock-on to a perspective that is dissimilar to all other perspectives but uninformative.
    \item It accounts for uncertainty in the estimated feature expectations by down-weighting old trajectories.
\end{enumerate*}

\begin{algorithm2e}[t]
\caption{Dissimilarity Sampling}\label{alg:corr_diag_selection}
\SetAlgoLined
\DontPrintSemicolon
\KwIn{perspectives $\perspectives$, approximate feature expectations $\hat{\matvec{\mu}} (\perspective)$, learner's policy $\pi^\learner$, similarity function $f$}

 \uIf{$\exists \perspective_i \in \perspectives: n(\perspective_i) = 0$}{
    \tcc{Perspective has not been selected yet}
    $\perspective \leftarrow \perspective_i$ \;
  }
\Else{
    \tcc{Calculate selection probabilities based on inverse similarities}
    \For{$i=1,\ldots, | \perspectives |$}{
        $s_i = 1 / \sum_{j \neq i} f (\hat{\matvec{\mu}} (\perspective_i), \hat{\matvec{\mu}}( \perspective_j)) \leftarrow$ Score for persp.\ $\perspective_i$ \;
        $p_i = s_i / \sum_j s_j \leftarrow$ Selection probability for $\perspective_i$
    }
    $\perspective \sim \textnormal{Categorical}([1, \ldots, |\perspectives|], [ p_1, \ldots, p_{| \perspectives |} ])$ \;
}
$\hat{\matvec{\mu}}' (\perspective) =  \sum_{t=0}^\infty \gamma^t \matvec{\phi}_\perspective (s_t,a_t) \leftarrow$ Estimate discounted feature expectations with rollout from $\pi^\learner$ in $\perspective_t$ \;
$\hat{\matvec{\mu}} (\perspective) = \alpha \hat{\matvec{\mu}} (\perspective) + (1 - \alpha) \hat{\matvec{\mu}}' (\perspective) \leftarrow$ Update exponentially weighted average of $\perspective$'s feature expectations

\KwResult{Next perspective $\perspective$} %
\end{algorithm2e}

%% file: a12_gail_background.tex
\section{Background: Generative Adversarial Imitation Learning}
\label{sec:gail_background}

Our approach builds on \emph{generative adversarial imitation learning} (GAIL)~\cite{ho2016generative}, which utilizes a GAN\cite{NIPS2014_5ca3e9b1}-inspired framework for training the learner's policy to mimic the expert's policy.

In GAIL, we use (given) expert behavior as demonstrations to train a learner policy without direct access to the reward signal or interaction with the expert.
Concretely, a discriminator is trained to distinguish between data from the expert and the learner. The learner aims to prevent the discriminator from distinguishing between expert data and its trajectories. An oracle discriminator would assign probability $1$ to state-action pairs generated by the learner and $0$ to the expert's state-action pairs \footnote{Note that perfect separation of the learner's and expert's state-actions might not be possible.}. The learner's goal is to find a policy $\pi^\learner_\theta$ such that the performance of the discriminator is minimized. 
Formally, this results in the following minimax optimization problem:
\begin{align}\label{eq:gail}
    \min_{\pi^\learner_\theta} \underbrace{\max_{\discriminator} \text{ } \mathbb{E}_{\pi^\learner_\theta} [\log (\discriminator(\bm s, \bm a))] + \mathbb{E}_{\pi^\expert} [1-\log (\discriminator(\bm s, \bm a))]}_{\text{Discriminator objective}} \quad \underbrace{\vphantom{\max_{\discriminator}} - \lambda H(\pi^\learner_\theta)}_{\mathclap{\text{entropy regularization}}}~,
\end{align}
where the discriminator $\discriminator\colon \states \times \actions \rightarrow [0,1]$ outputs the probability that the learner generated a state-action pair, $\mathbb{E}_{\pi}[\cdot]$ refers to the expectation with respect to state-action pairs observed when following some policy $\pi$, and where $\pi^\learner_\theta$ is the learner's policy parameterized by $\theta$.

The optimal solution to the above problem occurs when the learner's policy generates data that can not be distinguished from the expert's data by the discriminator.
In this case, for an arbitrary powerful discriminator, the learner's policy equals the expert's policy.

In practice, the discriminator and the learner's policy are alternatingly updated using stochastic gradient descent~\cite{ho2016generative}. 
The discriminator minimizes the negative binary cross-entropy, while the learner optimizes its policy using a (reformulated) output of the discriminator as a reward signal (e.g., using Trust Region Policy Optimization (TRPO)~\citep{pmlr-v37-schulman15} or Proximal Policy Optimization (PPO)~\citep{schulman2017ppo}).

In the case of not having direct access to actions but only observations (i.e., the case of learning from observations (LfO)), we reformulate the above equation based on \citet{torabi_generative_2019}:
\begin{align}\label{eq:gail_lfo}
    \min_{\pi^\learner_\theta} \max_{\discriminator} \; & \mathbb{E}_{\pi^\learner_\theta} [\log (\discriminator(\bm o_t, \bm o_{t+\Delta}))] + \mathbb{E}_{\pi^\expert} [1-\log (\discriminator(\bm o_t, \bm o_{t+\Delta}))] \\ & - \lambda H(\pi^\learner_\theta)~,
\end{align}
 where the discriminator $\discriminator$ is now a function $\discriminator\colon \states \times \states \rightarrow [0,1]$ mapping two observations with some specific delay $\Delta$ to the probability of how likely the tuple of states was generated from the learner.

%% file: a13_extended_related_work.tex
\section{Additional Related Work}
\label{sec:extended-related}

Several existing works consider active learning in the context of IL or IRL.
They share the goal of reducing the expert's effort and/or the number of interactions with the environment.
However, to the best of our knowledge, none of these considers the setting of our paper in which the learner must actively decide on the perspective from which it observes the expert.
They mainly focus on querying additional information regarding optimal actions in relevant parts of the state space or in out-of-distribution settings, e.g., by directly querying for the optimal action in specific states (e.g.,~\cite{judah2012active,brantley2020active,hussein2018deep}) or by requesting additional demonstrations in cases of uncertainty about optimal behavior (e.g,~\cite{di2020safari,lindner2022active}.
These works are complementary to ours and combining their approaches with ours might be an interesting direction for future work.

More specifically, \citet{judah2012active} for instance considers the setting in which the learner can query the expert about the best action for a particular state which is selected based on previous queries and environment interactions.
A similar approach is taken in \citet{brantley2020active} in which additionally a noisy oracle (also termed \textit{noisy heuristic}, a classifier) is considered which predicts the probability that an expert would not be consistent with the noisy oracle, and the expert is only consulted if that probability is sufficiently large. 
Thereby, the number of queries to the expert is minimized.
\citet{hussein2018deep} implement active IL to improve generalization in deep RL, querying actions for states for which the policy is uncertain, and, thereby, speeding up the learning process.
\citet{di2020safari} consider a setting in which the learner can request additional demonstrations in out-of-distribution settings and demonstrate that this can improve performance on manipulation tasks.

The recent survey of work on interactive imitation in robotics research by \citet{ROB-072} provides an overview of different possible query modalities and interfaces for human-robot interaction.

%% file: a14_compute_parameters.tex
\section{Architecture, Hyperparameters and Compute}\label{app:hyperparams_network_compute}
In the following section, we describe the architecture of our discriminators and the hyperparameters we used to get our results. Section~\ref{app:hyperparams_network_compute:compute} outlines our experiments' computational resources and hardware specifications.

\subsection{Network architectures}\label{app:hyperparams_network_compute:network}
Similar to \cite{torabi_generative_2019}, we concatenate RGB-images (in our case 2) of size $3  \times d \times d$ and feed them into our network as $6 \times d \times d$ arrays. We use image shifts ($\Delta>1$ as in \citet{stadie2017third}) instead of consecutive observations. Refer to the tables in  Section~\ref{app:hyperparams_network_compute:hyperparamaters} for the specific shifts used for each environment.
As we assume to be given only observations of the environment (i.e. images in our experiments, cf.\ Section~\ref{sec:experiments}) and do not observe the expert's actions, we build on Equation~\ref{eq:gail_lfo} substituting observations for states. Concretely, we pass a pair of observations through a convolutional feature extractor before using an MLP classification head with the Sigmoid activation function. We regularize all discriminators with Spectral Normalization \cite{miyato2018spectral}, which we found improves training stability. When learning correlations with multiple discriminators, we insert a 2 layer MLP network with shared parameters across all discriminators before a linear classification layer. In addition to the features extracted from the convolutional network, the first layer of the correlation network also conditions on the current perspective through a one-hot encoding. In both cases, we can interpret the output of the Sigmoid as the probability that the concatenated input observations belong to the learner's state distribution. The conditional discriminator concatenates the perspective information with the flattened features obtained after its convolutional layers. For Point, this information is the index of a perspective whereas for Reacher we use the normalized camera rotation parameters (azimuth, elevation, distance). The FiLM discriminator uses the same perspective information to generate the parameters for the FiLM blocks.

\begin{table}[htp]
\centering
\caption{Discriminator architecture for Point\\ with 1D observations.}
\label{tab:architecture:point_1D}
\centering
{
    \begin{tabular}{ l c  }
    \toprule
    \textbf{Hyperparameter } & \textbf{Value}  \\
    \midrule
     Observation rendering & (32, 32), RGB \\
     Image augmentation & None \\
     Spectral normalization & Yes \\
     \textsc{CNN Trunk} & \\
     \hspace{0.2cm} Convolution type & 1D \\
     \hspace{0.2cm} \# Convolution layers & 3 \\
     \hspace{0.2cm} Convolution filters & [16, 32, 32] \\
     \hspace{0.2cm} Convolution filter sizes & [4, 4, 4] \\
     \hspace{0.2cm} Convolution strides & [2, 2, 2] \\
     \hspace{0.2cm} Convolution padding & [0, 1, 0] \\
     \hspace{0.2cm} BatchNorm after Conv & [2, 3] \\
     \hspace{0.2cm} Activation function & LeakyReLU \\
     \textsc{Classification head} & \\
     \hspace{0.2cm} \# Linear layers & 2 \\
     \hspace{0.2cm} \# Hidden units & 32 \\
     \hspace{0.2cm} Activation function & LeakyReLU\\
     &  + Sigmoid\\
     \textsc{Correlation layer} & \\
     \hspace{0.2cm} \# Linear layers & 2 \\
     \hspace{0.2cm} \# Hidden units & 64 \\
     \hspace{0.2cm} Activation function & LeakyReLU\\
    \bottomrule
    \end{tabular}
}
\end{table}%

\begin{table}[htp]
\caption{Discriminator architecture for Reacher.}
\label{tab:architecture:mujoco}
\centering
    \begin{tabular}{ l c  }
    \toprule
    \textbf{Hyperparameter } & \textbf{Value}  \\
    \midrule
     Observation rendering & (64, 64), RGB \\
     Image augmentation & Flip \\
     Spectral normalization & No \\
     \textsc{CNN Trunk} & \\
     \hspace{0.2cm} Convolution type & 2D \\
     \hspace{0.2cm} \# Convolutional layers & 4 \\
     \hspace{0.2cm} Convolution filters & [16, 32, 32, 32] \\
     \hspace{0.2cm} Convolution filter sizes & [4, 4, 4, 4] \\
     \hspace{0.2cm} Convolution strides & [2, 2, 2, 4] \\
     \hspace{0.2cm} Convolution padding & [1, 1, 1, 0] \\
     \hspace{0.2cm} Activation function & LeakyReLU \\
     \textsc{Classification head} & \\
     \hspace{0.2cm} \# Linear layers & 2 \\
     \hspace{0.2cm} \# Hidden units & 64 \\
     \hspace{0.2cm} Activation function & LeakyReLU\\
     &  + Sigmoid\\
     \textsc{Correlation layer} & \\
     \hspace{0.2cm} \# Linear layers & 2 \\
     \hspace{0.2cm} \# Hidden units & 64 \\
     \hspace{0.2cm} Activation function & LeakyReLU\\
    \bottomrule
    \end{tabular}
\end{table}

We use binary cross entropy $\CE$ as an objective for optimizing our discriminators:
\begin{align*}
  \CE(\hat{y}_t, y_t) = -\big(y_t \log(\hat{y}_t) + (1-y_t) \log(1-\hat{y}_t)\big), %
\end{align*}
where $\hat{y}_t = \discriminator(\bm o_t, \bm o_{t+\Delta})$ is the discriminator's output. The discriminator's prediction target $y_t$ is $0$ in case the observation is from the expert and $1$ if it is from the learner. We use scaled uniform noise around the target labels to stabilize discriminator training.

\subsection{Computational Resources}\label{app:hyperparams_network_compute:compute}
Our experiments are generated using two servers with $2 \times$ 2x Intel 4214R CPUs each. Both machines are equipped with two Tesla A100 GPUs.
For hyperparameter tuning, we rely on sensible default values and use a simple grid search to tune the most important hyperparameters around those values totaling $620$ runs for Point and $700$ for Reacher.
The time required to run a single experiment depends on other variables such as current server load but is approximately $70$ minutes for Point and $30$ minutes for Reacher.

\begin{table}[htp]
\caption{Hyperparameters tuned using grid search.}
\label{tab:parameters:grid_search_params}
\centering
    \begin{tabular}{ l c  }
    \toprule
    \textbf{Hyperparameter} & \textbf{Grid Search Values}  \\
    \midrule
    Learning rate policy & $\{ 0.0001, 0.00025, 0.0005, 0.001 \}$ \\
    Learning rate value function & $\{ 0.0001, 0.00025, 0.0005, 0.001 \}$ \\
    Agent batch size & Subset of $\{32, 64, 128, 256, 512\}$ \\
    & depending on the algorithm  \\
    Discriminator learning rate & $\{ 0.0001, 0.00025, 0.0005, 0.001 \}$ \\
    \bottomrule
    \end{tabular}
\end{table}

\subsection{Hyperparameters}\label{app:hyperparams_network_compute:hyperparamaters}

\begin{table}[htp]
\caption{Hyperparameters for Point.}
\label{tab:parameters:point}
\centering
{
    \begin{tabular}{ l c  }
    \toprule
    \textbf{Hyperparameter } & \textbf{Value}  \\
    \midrule
    \# of seeds & 20\\
    \# of iterations & 40\\
    \textsc{PPO} & \\
    \hspace{0.2cm} Hidden size & [64, 64] \\
    \hspace{0.2cm} Trajectories per iter & 15 \\
    \hspace{0.2cm} Train episodes per iter & 100 \\
    \hspace{0.2cm} Optimizer & Adam\\
    \hspace{0.2cm} Learning rate policy & 0.00004 \\
    \hspace{0.2cm} Learning rate value function & 0.0001 \\
    \hspace{0.2cm} Batch size & 512 \\
    \hspace{0.2cm} GAE $\lambda$ & 0.9995 \\
    \hspace{0.2cm} Entropy coefficient $\alpha$ & 0.0 \\
    \textsc{Discriminators} & \\
    \hspace{0.2cm} Trajectories per iter & 50\\
    \hspace{0.2cm} Train epochs & 10\\
    \hspace{0.2cm} Optimizer & Adam\\
    \hspace{0.2cm} Learning rate & 0.0007 \\
    \hspace{0.2cm} Batch size & 64 \\
    \hspace{0.2cm} Image shift & 2\\
    \textsc{Environment} &  \\
    \hspace{0.2cm} Discount factor $\gamma$ & 0.99 \\
    \hspace{0.2cm} Maximum episode length & 42 \\
    \hspace{0.2cm} Observation rendering & (32, 32), RGB \\
    \bottomrule
    \end{tabular}
}
\end{table}

\begin{table}[htp]
\caption{Hyperparameters for Reacher.}
\label{tab:parameters:reacher}
\centering
{
    \begin{tabular}{ l c  }
    \toprule
    \textbf{Hyperparameter } & \textbf{Value}  \\
    \midrule
    \# of seeds & 20\\
    \# of iterations & 40\\
    \textsc{PPO} & \\
    \hspace{0.2cm} Hidden size & [32, 32] \\
    \hspace{0.2cm} Trajectories per iter & 15 \\
    \hspace{0.2cm} Train episodes per iter & 10 \\
    \hspace{0.2cm} Optimizer & Adam\\
    \hspace{0.2cm} Learning rate policy & 0.00095 \\
    \hspace{0.2cm} Learning rate value function & 0.0002 \\
    \hspace{0.2cm} Batch size & 256 \\
    \hspace{0.2cm} GAE $\lambda$ & 0.97 \\
    \hspace{0.2cm} Entropy coefficient $\alpha$ & 0.003 \\
    \textsc{Discriminators} & \\
    \hspace{0.2cm} Trajectories per iter & 10 \\
    \hspace{0.2cm} Train epochs & 10 \\
    \hspace{0.2cm} Optimizer & Adam\\
    \hspace{0.2cm} Learning rate & 0.0002 \\
    \hspace{0.2cm} Batch size & 128 \\
    \hspace{0.2cm} Image shift & 5 \\
    \textsc{Environment} &  \\
    \hspace{0.2cm} Discount factor $\gamma$ & 0.99 \\
    \hspace{0.2cm} Maximum episode length & 54 \\
    \hspace{0.2cm} Observation rendering & (64, 64), RGB \\
    \bottomrule
    \end{tabular}
}
\end{table}

%% file: a18_illustration_thm.tex
\section{Empirical Validation of Theorem~\ref{thm:linear-rewards-and-projections}}
\label{sec:validation+lp}

In this section, we empirically validate Theorem~\ref{thm:linear-rewards-and-projections}.

\textbf{Environment.} To this end, we consider grid worlds of size $10 \times 10$ in which an agent can move up, down, left, or right.
In each instance of the grid world, there are $8$ objects of $k=4$ different types ($2$ objects of each type) distributed uniformly at random across the grid world's cells.
Each object type is associated with a random non-negative reward sampled uniformly from $[0,1]$, i.e., the reward of the $i$\textsuperscript{th} object type is $w_i^* \sim \mathcal{U}([0,1])$.
When reaching a grid cell with an object in it, the agent receives the corresponding reward upon moving out of the cell and is randomly placed in a cell without an object.
Note that the rewards described above are linear in state-dependent features which correspond to indicator vectors representing the type of object present in a state, i.e., $r(s,a) = \langle \phi(s,a), [w_1^*, \ldots, w_k^*]^T \rangle$,
where $\phi(s,a) = [\mathbf{1}_\textnormal{object of type 1 is in state $s$}, \ldots, \mathbf{1}_\textnormal{object of type $k$ is in state $s$}]^T$.
We consider a continuing setting (i.e., an infinite horizon), where the agent's starting position is a randomly selected empty cell.
For computing cumulative rewards, we use a discount factor $\gamma=0.3$.

\textbf{Learner.}
The learning agent does not observe the features described above directly. Instead, it views them through perspectives corresponding to linear transformations.
In particular, we consider the following two settings:
\begin{itemize}
    \item \textsc{Subset}: The agent observes the first $1 \leq i \leq k$ features, i.e., $\perspectives=\{A_1, \ldots, A_i\}$, where $A_i \in \mathbb{R}^{1 \times k}$ such that
    \begin{align*}
        A_i = \begin{bmatrix}
            \mathbf{1}_\textnormal{$i=1$}, \mathbf{1}_\textnormal{$i=2$}, \ldots, \mathbf{1}_\textnormal{$i=k$}
          \end{bmatrix}.
    \end{align*}

    \item \textsc{Random}: The agent observes $1 \leq i \leq k$ projections of the features onto a random vector, i.e., $\perspectives=\{A_1, \ldots, A_i\}$, where $A_i \in \mathbb{R}^{1 \times k}$ such that
    \begin{align*}
        A_i = \begin{bmatrix}
            A_{i,1}, A_{i,2}, \ldots, A_{i,k}
          \end{bmatrix},
    \end{align*}
    where $A_{i,j} \sim \mathcal{U}([0,1])$ for $j \in [k]$.

\end{itemize}

\textbf{Algorithms, the expert, and learning.}
We obtain an expert policy for a grid world instance through linear programming (LP) and use it to generate demonstrations.
The learner aims to match the empirical feature frequencies of the expert in all perspectives available to it.
We perform feature matching using an LP formulation minimizing the $\ell_\infty$ distance between the empirical feature expectations computed from the demonstrations and the feature expectations realized by the learner (viewed through the perspectives available to the learner).
In particular, given demonstrations $\tau_1, \ldots, \tau_T$ in perspectives $A_{j_1}, \ldots A_{j_T}$, where $j_t$ is the index of the perspective used at time $t$, the learner solves the following LP:
\begin{align}
    \min_{\mu, \epsilon_1, \ldots, \epsilon_{|\perspectives|}} \quad& \sum_{i=1}^{|\perspectives|} \epsilon_i \\
    \textnormal{s.t.} \quad & \sum_a \mu(s,a) - \gamma \sum_{s'} \sum_a \mu(s',a) T(s' | s, a) = \rho(s) \\
      &\qquad\qquad \forall s, a \quad \textnormal{(Bellman flow equations)} \nonumber \\
      & \| A_i F \mu - \hat{\Psi}_i \|_\infty \leq \epsilon_i \\
      &\qquad\qquad \forall i \in \perspectives \quad \textnormal{(feature matching)} \nonumber
\end{align}
where $\mu$ is the so-called state occupancy, $\rho(s)$ is the initial state probability, $F \in \mathbb{R}^{k \times (|\states \cdot \actions|)}$ is a mapping assigning each state-action-pair a feature vector in $\mathbb{R}^k$, and where $\hat{\Psi}_i$ is the empirical feature expectation from the demonstrations in the $i$th perspective.

\begin{figure*}
  \subcaptionbox{Subset of features (\textsc{Subset}) \label{fig:thm:illustration:fixed}}[.5\linewidth]{%
    \includegraphics[]{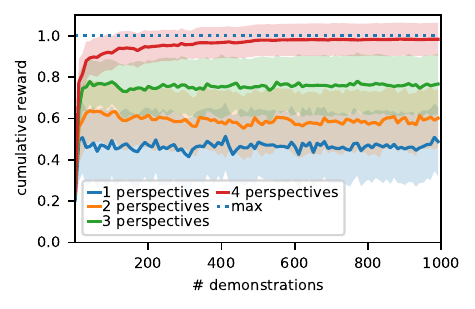}%
  }%
  \hfill
  \subcaptionbox{Random linear transformations (\textsc{Random}) \label{fig:thm:illustration:random}}[.5\linewidth]{%
    \includegraphics[]{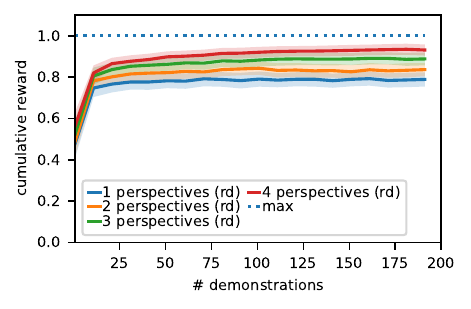}%
  }
    \caption{Learner's performance for an increasing number of expert demonstrations.
    We show expected discounted cumulative rewards, normalized so that the maximum achievable reward is $1$.
    Available perspectives: \textit{(\subref{fig:thm:illustration:fixed})} \textsc{Subset} \textit{(\subref{fig:thm:illustration:random})} \textsc{Random}. For 4 available perspectives, all features (or linear transformations of those with joint full rank (whp)) can be observed and optimal performance is approached with an increasing number of demonstrations. For fewer perspectives, the performance approaches and converges to a level below the maximum achievable performance.   
    }
  \label{fig:thm:illustration}
\end{figure*}

\textbf{Results.}
In Figure~\ref{fig:thm:illustration}, we show the performance of the learner for different numbers of available perspectives and increasing numbers of expert demonstrations. Results are averaged over 10 random grid worlds according to the above description.
We observe that for an increasing number of demonstrations the performance of all agents endowed with any number of perspectives improves.
For sufficiently many perspectives, the performance converges to the optimal achievable performance (computed using the true reward function).
Comparing \textsc{Subset} and \textsc{Random}, for the same number of available perspectives, better performance is achieved using the random projections, likely because of the random features combining information about multiple ground truth features.

\section{Grid worlds for Evaluating Active Perspective Selection Strategies}
\label{sec:gws} 

In this section, we provide additional details regarding the experiments presented in Section~\ref{sec:feasibility}.

\paragraph{Environment.}
We consider grid worlds of size $10 \times 10$ in which 2 objects of 4 different types are placed randomly upon creation of the environment.
Each object type is assigned a reward randomly drawn from $\mathcal{U}([0,1])$.
An agent receives an object's reward upon collecting it by moving on the respective cell.
The agent can move deterministically along the cardinal directions with the exception of collecting an object in which case the agent is randomly moved to an empty cell.
The ground-truth features of the environment are indicators for the 4 different object types, i.e., $\phi(s,a) \in \mathbb{R}^4$.
The interaction of the agent with the environment is non-episodic, and we use a discount factor $\gamma=0.3$.

\paragraph{Perspectives.}
\begin{itemize}
    \item For the basis-vector transformations we consider the following 16 perspectives given by their linear transformation matrices $A_i$, $i=1, \ldots, 16$:
      \begin{align*}
          A_1 &= [1,0,0,0] \\
          A_2 &= [0,1,0,0] \\
          A_3 &= [0,0,1,0] \\
          A_4 &= [0,0,0,1] \\
          A_i &= [1,0,0,0] \quad i \in {5, \ldots, 16} \\
      \end{align*}
      That is, the perspectives $A_1, A_5, \ldots, A_{16}$ are the same.

    \item For the random linear transformations 40 perspectives given by their linear transformation matrices $A_i$, $i=1, \ldots, 40$, are created as follows:
      \begin{enumerate}
          \item Draw $\tilde{A}_i \in \mathbb{R}^4$ from $\mathcal{U}([0,1]^4)$.
          \item Construct $A_i$ from $\tilde{A}_i$ by setting all entries below $0.5$ to zero. This ensures that each perspective only measures a subset of the ground truth features (in expectation).
      \end{enumerate}
      Because of the large number of perspectives, it is probable that some of the perspectives are similar, i.e., contain \emph{redundant} information.
\end{itemize}

\paragraph{Perspective selection strategies.}
We consider 4 perspective selection strategies:
\begin{itemize}
    \item \textsc{uniform}. This strategy selects the available perspectives in a round-robin fashion, i.e., in the $j$th interaction it selects the perspective $j \mod K$, where $K$ is the total number of available perspectives.

    \item \textsc{active (var)}. This strategy exploits full knowledge about the feature transformations. In particular, the feature matching problem is considered as a least-squares problem in which the matrices $A_i$ correspond to the regressors, and the observed feature expectations for a demonstration correspond to the independent variable.
    In this setting, one can compute the variance of the estimate of the optimal regression coefficients. 
    The perspectives are selected in order to minimize the variance of this estimate.

    \item \textsc{active (sim)}. This strategy exploits some knowledge about the feature transformations provided in terms of similarities among perspectives.
    Concretely, the inner product of pairs of feature transformations (normalized by the 2-norm of the transformations) is considered as their similarities and used to construct a similarity matrix $S$. 
    From this similarity, we compute probabilities for sampling each perspective by considering the normalized inverse similarity of the perspective to all other perspectives.

    \item \textsc{active (corr)}. 
      This strategy is similar to \textsc{active (sim)} but does not leverage any knowledge about the feature transformations. Similarities are replaced by feature correlations among different perspectives similar as detailed in Section~\ref{app:corr_diag_selection}.
    
\end{itemize}